\documentclass{SCIS2026}
\usepackage{url}            
\usepackage{booktabs}       
\usepackage{amsfonts}       
\usepackage{nicefrac}       
\usepackage{microtype}      
\usepackage{xcolor}         

\usepackage{caption}
\usepackage{amsmath}
\usepackage{amssymb}
\usepackage{mathtools}
\usepackage{amsthm}
\usepackage{bm}
\usepackage{diagbox}
\usepackage{makecell}
\theoremstyle{plain}
\theoremstyle{remark}
\usepackage{tcolorbox}
\usepackage{subcaption}

\usepackage{wrapfig}
\usepackage{multirow}

\newcommand{\RightComment}[1]{\hfill{\scriptsize\textcolor{gray}{$\triangleright$ #1}}}

\begin{document}
\ArticleType{LETTER}
\Year{2025}
\Month{January}
\Vol{68}
\No{1}
\DOI{}
\ArtNo{}
\ReceiveDate{}
\ReviseDate{}
\AcceptDate{}
\OnlineDate{}
\AuthorMark{}
\AuthorCitation{}

\title{Stop Wandering, Find the Keys: LLMs Discriminate Key States for Efficient Multi-Agent Exploration}{Stop Wandering, Find the Keys: LLMs Discriminate Key States for Efficient Multi-Agent Exploration}

\author[1]{Yun QU}{}
\author[1]{Boyuan WANG}{}
\author[1]{Yuhang JIANG}{}
\author[1]{Jianzhun SHAO}{}
\author[1]{Yixiu MAO}{}
\author[1]{\\Heming ZOU}{}
\author[1]{Chang LIU}{}
\author[1]{Qi WANG}{cheemswang@mail.tsinghua.edu.cn}
\author[2,3]{Meiqin LIU}{liumeiqin@zju.edu.cn}
\author[1]{Xiangyang JI}{xyji@tsinghua.edu.cn}

\address[1]{Department of Automation, Tsinghua University, Beijing 100084, China}
\address[2]{College of Electrical Engineering, Zhejiang University, Hangzhou 310027, China}
\address[3]{National Key Laboratory of Human-Machine Hybrid Augmented Intelligence, Xi’an Jiaotong University, Xi’an 710049, China}

\maketitle

\begin{multicols}{2}

\begin{figure*}
    \centering
    \vspace{-10pt}
    \includegraphics[width=0.95\linewidth]{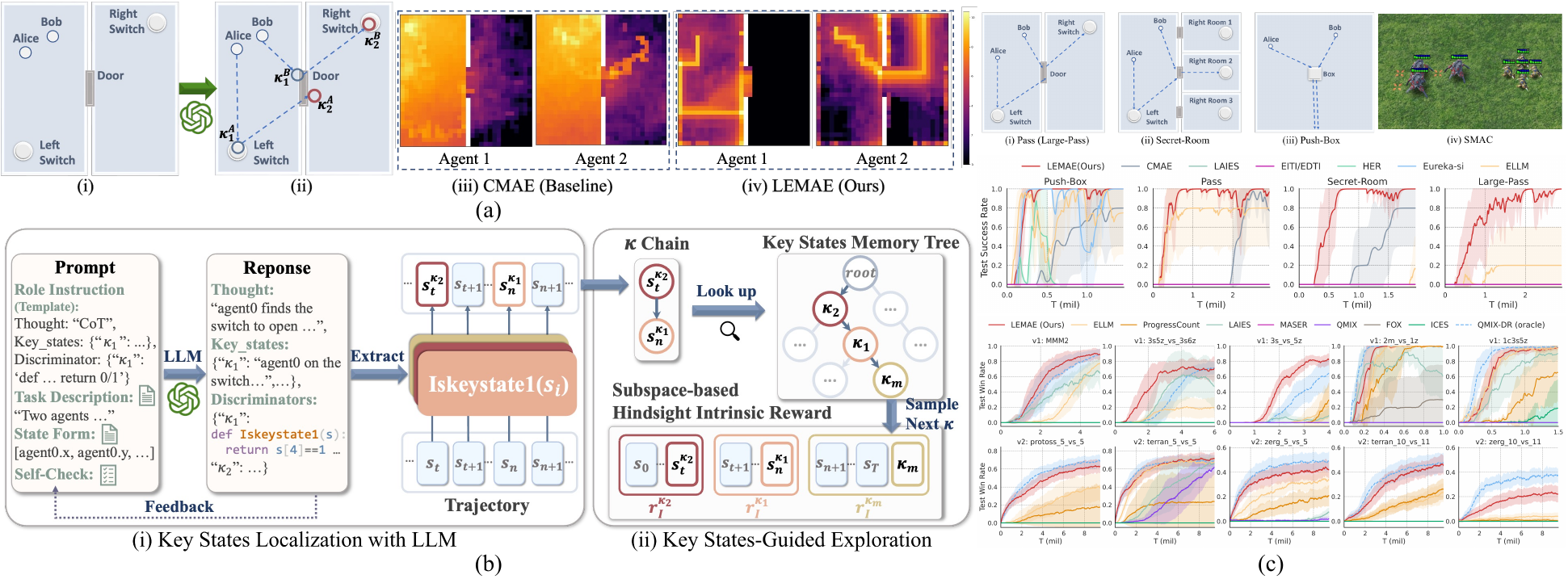}
    \vspace{-8pt}
    \caption{
Overview and evaluation of LEMAE. 
(a) LLM-annotated key states, agent visitation maps for CMAE (baseline) and LEMAE, showing reduced redundant exploration and emergent division of labor. 
(b) LEMAE pipeline: key states localization with LLM and key states-guided exploration. 
(c) Evaluation on MPE and SMAC tasks under sparse rewards, demonstrating consistent improvement over baselines.
}
    \vspace{-10pt}
    \label{fig:scis}
\end{figure*}
\noindent Exploration is a fundamental issue in reinforcement learning~(RL). 
Researchers have developed several exploration strategies directed by novelty, diversity, or uncertainty, mainly in single-agent RL. 
While these methods have made great progress, they may induce task-irrelevant redundant exploration in complex environments~\cite{du2023guiding}. 
Especially in multi-agent reinforcement learning~(MARL), the need to mitigate exploration redundancy becomes even more pressing due to the challenges like exponential expansion of the joint state-action space. 
Widespread real-world applications, including MOBA games, social science, and multi-vehicle control, further underscore the urgency for efficient multi-agent exploration.

This work introduces a new direction: leveraging \textit{task-relevant guidance} as a key to facilitate more efficient cooperative multi-agent exploration.
Incorporating structured priors in exploration typically requires domain expertise and significant manual effort.
Hopefully, recent advances have witnessed the remarkable reasoning and planning capabilities of large language models (LLMs)
, offering a promising source of generalizable prior knowledge to enhance efficient exploration.
However, incorporating linguistic LLM priors into symbolically represented RL tasks remains non-trivial~\cite{du2023guiding}.
Bridging this semantic-symbolic gap in a simple and lightweight manner is thus crucial for practical deployment.

To address this, we propose LEMAE, a novel framework for efficient and coordinated multi-agent exploration via LLM-derived guidance.
LEMAE comprises two core components: 
(i) \textit{key states localization with LLM}, and (ii) \textit{key state-guided exploration}.
The \textit{first component} automatically localizes task-critical key states, through LLM discrimination, thus injecting LLM priors into the RL process.
Specifically, the LLM-induced discriminator functions work to discriminate key states from rollout trajectories, enhancing reliability while reducing the overburden of LLM inference costs.
The \textit{second component} uses these localized key states to guide exploration.
Concretely, key states are treated as explicit training signals, and we introduce Subspace-based Hindsight Intrinsic Reward (SHIR) to densify rewards for achieving guided exploration.
To further organize exploration with memory, we construct a Key States Memory Tree (KSMT) to track key state transitions, simplify the exploration space, and enhance the guidance effect of SHIR.
As illustrated in the heatmap in Figure~\ref{fig:scis}(a), LEMAE achieves a significant performance advantage through notably reducing redundant exploration and fostering organic cooperation among agents.

Our \textbf{main contributions} are summarized as follows: 

\begin{enumerate}
\vspace{-8pt}
    \item We build a bridge between LLM and RL to facilitate efficient multi-agent exploration by developing a systematic approach dubbed LEMAE.
    \vspace{-8pt}
    \item We devise a computationally efficient inference strategy channeling task-specific information from LLM to discriminate task-crucial key states, which are used as subgoals with agent-wise labor division for coordinated 
    and targeted exploration. 
    \vspace{-8pt}
    \item We introduce Key State Memory Tree to organize exploration according to historic key state transitions, and devise the Subspace-based Hindsight Intrinsic Reward to promote guided exploration.
    \vspace{-8pt}
\end{enumerate}

\lettersection{Methods}
Prior task-agnostic novelty-driven strategies~\cite{du2023guiding} may induce redundant exploration, especially in MARL where the joint state-action space scales exponentially with the number of agents.
We introduce Key States as task-relevant priors to guide efficient multi-agent exploration.
Proposition~\ref{randomwalk_lemma} formalizes their efficiency advantage over unguided wandering (proof in Appendix).

\definition[Key States]{Key States are task-critical intermediate states that (i) admit explicit representations and (ii) capture the functional labor division among agents in multi-agent cooperation.}

\proposition{Consider a one-dimensional asymmetric random walk starting at $0$ and targeting $N>1$, with right-move probability $p\in(0.5,1)$.
Without prior knowledge, the expected first hitting time is $\mathbb{E}(T_{0\rightarrow N}) = \frac{N}{2p-1}$.
After introducing the task-relevant information that the agent must first reach key states $\kappa=1, \dots, N-1$ before reaching $x=N$, the hitting time decreases $\mathbb{E}(T_{0\rightarrow N}) - \mathbb{E}(T^{\mathrm{prior}}_{0\rightarrow N}) = (N-1)*(\frac{1}{2p-1} - \frac{2}{p}+1) > 0$.  
\label{randomwalk_lemma}}

This indicates that Key States implicitly decompose global exploration into structured subgoal-directed transitions, reducing complexity and improving sample efficiency. The intuition aligns with practical mechanisms such as checkpoints in games.

To reduce manual effort, LEMAE employs LLMs in a discriminative manner to localize key states from rollout trajectories, which is more reliable in complex MARL.
For each key state $\kappa_i$, LLM generates a corresponding discriminator function $\mathcal{F}_i : \mathcal{S} \rightarrow \{0,1\}$ to annotate each state, 
where $\mathcal{S}$ denotes the state space and $\mathcal{F}_i(s_t)=1$ indicates that state $s_t$ satisfies the semantic criteria of $\kappa_i$.
A \textit{standardized prompt template} with role instructions, output constraints, and labor-division guidance requires
only \textit{minimal task details, the description and the state form}, which can be easily extracted from the document.
A Self-Check mechanism (LLM rethinking and code verification) mitigates hallucinations and ensures executable functions.
This design incurs low overhead (typically $\le3$ LLM calls per task) and avoids reliance on environment code.

Key states act as explicit signals that densify sparse rewards and segment trajectories into subgoal phases. 
Using the discriminator-induced reward-related subspace and inspired by HER~\cite{andrychowicz2017hindsight}, we define the Subspace-based Hindsight Intrinsic Reward (SHIR):
\vspace{-4pt}
\begin{equation}
\vspace{-2pt}
r_I^{\kappa_m}(t)=\|\Phi_m(s_t)-\Phi_m(\kappa_m)\|
-\|\Phi_m(s_{t+1})-\Phi_m(\kappa_m)\|,
\end{equation}
where $\Phi_m(s)=(s_e)_{e\in\upsilon_m}$ restricts the state to the discriminator-relevant subspace $\upsilon_m$. 
Operating in the subspace mitigates redundancy and bias from full-state reward shaping.
For time step $t$, let $m_t$ denote the index of the key state 
terminating the sub-trajectory to which $t$ belongs. 
Let $r_E(t)$ denotes the extrinsic environment reward. The overall reward is
\vspace{-3pt}
\begin{equation}
r(t)=\alpha \cdot r_E(t)
+\beta \cdot r_I^{\kappa_{m_t}}(t),
\qquad \alpha,\beta\ge0.
\end{equation}

To organize efficient exploration with memory, we introduce the Key States Memory Tree (KSMT), which models transitions among key states. 
KSMT is incrementally expanded using key state chains extracted from annotated trajectories until success transitions are sufficiently captured.
A mixed-randomness KSMT Explore strategy balances branch expansion and policy progression: at node $\xi_i$, high randomness is chosen with probability $p_i=\frac{1}{d_i+1}$ (where $d_i$ is node degree), and low randomness otherwise.
Branches not leading to success are pruned.
As a dynamic model over the key state space, KSMT supports subgoal planning by retrieving validated successors of the current key state chain. 
If none exist, a key state is sampled uniformly to maintain guidance. 
This mechanism enhances SHIR and improves exploration efficiency.

\lettersection{Experiments}
We conduct main experiments on typical multi-agent benchmarks: (1) the Multiple-Particle Environment (MPE) and (2) the StarCraft Multi-Agent Challenge (SMAC) v1 and v2~\cite{ellis2024smacv2}, as shown in Figure~\ref{fig:scis}(c).
We focus mainly on tasks with symbolic state spaces and sparse reward. 
We compare LEMAE with representative methods: (i) traditional methods for efficient multi-agent exploration and (ii) LLM-based methods with the same information.
We run with five random seeds.

On MPE tasks, LEMAE consistently outperforms all baselines, achieving up to 10$\times$ acceleration in discovering the first success state and significantly higher final success rates. 
Visitation analysis shows that LEMAE reduces redundant exploration and promotes emergent division of labor, while its pruning mechanism effectively mitigates task-irrelevant key states. 
On fully sparse SMAC v1/v2 tasks, LEMAE demonstrates clear superiority across diverse difficulty levels and agent scales, even matching or surpassing dense-reward oracles (QMIX-DR). 
Performance gains are more pronounced in harder and large-agent scenarios, highlighting the advantage of key state guidance in exponentially large spaces.

Removing SHIR or KSMT leads to substantial degradation, confirming their complementary roles. 
Combining strong LLMs with the Self-Check mechanism ensures high discriminator executability and stable performance. 
LEMAE is robust to reward scaling factors and maintains strong performance under simulated key-state perturbations (missing or distracting states), indicating tolerance to imperfect LLM outputs.
LEMAE generalizes to unseen tasks, extends to vision-based settings via multi-modal LLMs, and applies to continuous-control robotics benchmarks, where it approaches dense-reward performance under sparse supervision. 
It is algorithm-agnostic and consistently improves IPPO, QMIX, QPLEX, and VMIX, demonstrating broad compatibility.

\lettersection{Conclusion}
We present LEMAE, a novel framework that benefits multi-agent exploration with task-specific guidance from LLM.
LEMAE executes the \textit{key states localization with LLM} and enables the \textit{key state-guided exploration} to improve sample efficiency.
In this way, we can (i) establish connections between LLM and RL to ground linguistic knowledge into decision-making, (ii) reduce the manual workload in accessing knowledge and intensive inference calls from LLM, and (iii) significantly boost exploration efficiency through guided and organized exploration.
Extensive experiments further examine the effectiveness of LEMAE in typical benchmarks.

\Acknowledgements{This work is funded by National Natural Science Foundation of
China (NSFC) with the Number \# 62495091 and \# 62306326.}

\Supplements{Appendix A-J.}

\bibliographystyle{scis}
\bibliography{reference}





\end{multicols}

\newpage
\begin{appendix}
\section{Preliminary}
The environments considered in this work are characterized as a decentralized partially observable Markov decision process~(Dec-POMDP)~\cite{oliehoek2016concise} with $n$ agents, which can be defined as a tuple $G=\langle S,A,I,P,r,Z,O,n,\gamma \rangle$, where $s\in S$ is the global state, $A$ is the action space for each agent, and $\gamma \in [0, 1)$ is the discount factor. 
At time step $t$, each agent $i\in I\equiv\{1,...,n\}$ has its local observations $o^i\in O$ drawn from the observation function $Z(s, i):S\times I \rightarrow O$ and chooses an action $a^i\in A$ by its policy $\pi^i (a^i|o^i):O \rightarrow \Delta([0,1]^{|A|})$, forming a joint action $\mathbf{a}\in \mathbf{A}\equiv A^{n}$. 
$T(s'|s, \mathbf{a}):S\times\mathbf{A}\times S \rightarrow [0,1]$ is the environment's state transition distribution. 
All agents share a common reward function $r(s, \mathbf{a}):S\times\mathbf{A}\rightarrow \mathbb{R}$. 
The agents’ joint policy $\bm{\pi}:=\prod_{i=1}^n \pi^i$ induces a joint \emph{action-value function}: $Q^{\bm{\pi}}(s, \mathbf{a})=\mathbb{E}[R|s,\mathbf{a}]$, where $R=\sum^\infty_{t=0}\gamma^t r_{t}$ is the expected discounted return. The goal of MARL is to find the optimal joint policy $\bm{\pi}^*$ such that $Q^{\bm{\pi}^*}(s,\mathbf{a})\ge Q^{\bm{\pi}}(s, \mathbf{a})$, $\forall\bm{\pi}\ \text{and}\ (s,\mathbf{a})\in S\times \mathbf{A}$. Notably, we specifically focus on sparse reward tasks, i.e., $r_t=1$ only when $s_{t+1}=s_{success}$ where $s_{\text{success}}$ denote any terminal state satisfying the task success condition, otherwise $r_t=0$. 
We denote the symbol for the $i$-th key state by $\kappa_i$ together with its discriminator function $\mathcal{F}_i$.

\section{Related Works}
\textbf{LLM in Decision Making.}
Large language models have shown impressive capabilities across tasks~\cite{touvron2023llama}. 
Recent advances show a growing use of LLMs in decision-making.~\cite{wang2023survey}. 
A primary challenge is grounding LLM's linguistic knowledge into low-level control typically represented in symbolic form~\cite{peng2023self, carta2023grounding,qu2024latent}.
Some works use LLMs for high-level planning but they rely on low-level policies~\cite{yao2022react, shinn2023reflexion, zhu2023ghost, wang2023voyager} or APIs~\cite{liang2023code}, which limits their applicability.
Recently, LLMs have been integrated with RL to improve low-level control~\cite{cao2024survey}.
LLMs can be used as information processors~\cite{paischer2024semantic, kim2024efficient, wang2024llm,qu2024latent} or policy backbones~\cite{carta2023grounding, shi2023unleashing} to reduce learning complexity or improve performance, but cannot directly facilitate efficient exploration or demand enormous data and resources.
LLMs as goal selectors~\cite{su2023subgoal, Shukla2023lgts, zhuang2024yolo}, teacher policy~\cite{zhou2023large}, or task sampler~\cite{zhang2023omni} require predefined task pools, skills or subgoals.
LLM-based reward or policy design~\cite{klissarov2023motif, ma2023eureka,kwon2023reward, liu2024rl,chen2024rlingua} has made significant progress but often relies on annotated data, frequent LLM inferences, or abundant task information.
LLaMAC~\cite{zhang2023controlling} and ELLM~\cite{du2023guiding} use LLMs to enhance exploration, but they rely on predefined symbolic observation captioners and frequent LLM inferences.
In contrast, LEMAE integrates linguistic LLM priors into symbolic states with minimal task information and inference costs, achieved by localizing key states in trajectories using LLM-generated discriminator functions.

\textbf{Efficient Multi-Agent Exploration.}
Exploration efficiency has long been a focal point in RL~\cite{thrun1992efficient, cai2020provably, seo2021state,ecoffet2019go}.
Typical exploration methods focus on random exploration~\cite{mnih2013playing,rashid2018qmix}
or heuristic indicators, such as diversity or novelty, to facilitate exhaustive exploration, particularly in single agent exploration~\cite{linke2020adapting,  burda2018exploration, pathak2017curiosity, bellemare2016unifying}. 
Despite their success, they may induce notable redundant exploration due to a lack of task-relevant guidance~\cite{du2023guiding}.
The exponential expansion of the state-action spaces renders exhaustive exploration impractical in multi-agent settings. 
Consequently, efficient multi-agent exploration~(MAE) becomes increasingly imperative and necessary~\cite{jeon2022maser, liu2021cooperative, mahajan2019maven, feng2025multi}.
MAE is also challenging due to the complex configurations, e.g., the entangled effect of multi-agent actions and intricate reward design~\cite{liu2023lazy, xu2023subspace}.
Given our emphasis on efficient exploration, we prioritize evaluation in MAE.
Some MAE methods encourage influential behaviors during agent interactions~\cite{liu2023lazy, jaques2019social,wang2019influence}, but may cause unintended coalitions or need extra priors~\cite{liu2023lazy}.
Subgoal-based methods~\cite{jeon2022maser} face challenges in integrating task-related information into subgoals~\cite{tang2018hierarchical, kulkarni2016hierarchical,jeon2022maser,liu2021cooperative}.
WToE~\cite{dong2023wtoe} focuses on exploration timing but may face efficiency issues due to the absence of guided rewards.
To mitigate complexity, recent advances propose leveraging intrinsic structures, such as causal empowerment~\cite{cao2025causal1, cao2025causal2} or planning mechanisms~\cite{liu2024efficient}, to prioritize reward-relevant information. Similarly, FoX~\cite{jo2024fox} and ICES~\cite{li2024ices} utilize formation-based abstraction and latent Bayesian surprise to reduce the search space. 
However, these methods primarily rely on mining statistical or structural patterns from interaction data, often lacking the ability to incorporate explicit task-related priors, which results in potential redundant exploration before such structures are learned.
Distinguished from the above, this work underscores the significance of task-relevant guidance in exploration and utilizes the key state priors extracted from LLM to enable efficient multi-agent exploration.

\section{Method}

This section first induces the concept of key states as task-relevant guidance~(\ref{sec:ks}).
Centering around the key states, we construct two components: (i) \textit{key states localization with LLM}~(\ref{sec:LLMprompt}) and (ii) \textit{key state-guided exploration}~(\ref{sec:exploremethod}).
The former directs LLM to generate discriminator functions for localizing key states, while the latter guides exploration with the introduced Subspace-based Hindsight Intrinsic Reward and Key States Memory Tree.
Implementation details are in Figure~1(b) and Algorithm~\ref{alg:LEMAE}.
Meanwhile, we provide a \textcolor{cyan}{\href{https://sites.google.com/view/lemae}{demonstration}} to illustrate LEMAE's execution pipeline.

\subsection{Devil is in the Key States}\label{sec:ks}

As noted in the literature, previous methods suffer from redundant exploration in pursuit of task-agnostic novelty~\cite{du2023guiding}, potentially reducing training efficiency.
This issue becomes more pronounced in MARL, where the exploration space expands exponentially with agents due to interaction coupling and coordination demands.
This motivates us to integrate task-relevant information as key guidance for efficient exploration.
However, practical proposals are limited in the field. 
This work introduces \textbf{Key States} as a novel prior.
\definition[Key States]{Key States are task-critical intermediate states that (i) admit explicit representations and (ii) capture the functional labor division among agents in multi-agent cooperation.}

Proposition~1 highlights the efficacy of incorporating them instead of wandering.

\noindent\textbf{Proposition 1. }
Consider a one-dimensional asymmetric random walk starting at $0$ and targeting $N>1$, with right-move probability $p\in(0.5,1)$.
Without prior knowledge, the expected first hitting time is $\mathbb{E}(T_{0\rightarrow N}) = \frac{N}{2p-1}$.
After introducing the task-relevant information that the agent must first reach key states $\kappa=1, \dots, N-1$ before reaching $x=N$, the hitting time decreases $\mathbb{E}(T_{0\rightarrow N}) - \mathbb{E}(T^{\mathrm{prior}}_{0\rightarrow N}) = (N-1)*(\frac{1}{2p-1} - \frac{2}{p}+1) > 0$.  
\\

The proof is deferred to \ref{appsec:lemmaproof}.
The exploration policy substantially benefits from the involvement of key states, e.g., $\mathbb{E}(T_{0\rightarrow N}) - \mathbb{E}(T^{\mathrm{prior}}_{0\rightarrow N})\to\infty$ with $p\to 0.5$.
Intuitively, task-relevant key states implicitly induce a decomposition of the exploration space, which reduces exploration complexity from a global, undirected search to a series of subgoal-directed local transitions, guiding agents to explore along meaningful, task-relevant paths rather than random walks and improving sample efficiency, as illustrated in Figure~1(a).
Such a concept is also commonly seen in practical scenarios, such as in-game checkpoints~\cite{demaine2016super}
and landmarks in navigation~\cite{becker1995reliable}.

\subsection{Key States Localization with LLM}\label{sec:LLMprompt}
\vspace{-1pt}
To reduce manual workload, we employ LLM to localize key states.
Although generating key states can be straightforward, LLM's weakness in comprehending symbolic states or environment details necessitates additional information and can lead to errors and hallucinations difficult to detect.
We stress \textit{the importance of LLM's discriminative ability to localize key states in rollout trajectories} to better leverage its general knowledge.
Considering complex and partially observable MARL tasks, discrimination is more reliable and universal than generation, as discussed in \ref{discussion:discrimination}.

\textbf{Key state discriminator functions} $\{\mathcal{F}_i\}_{i=1}^m$ are generated by instructing LLM, as depicted in Figure~1(b), where $m$ is determined by LLM.
Each function $\mathcal{F}_i$, the \textit{Iskeystate$i$(s)} block, takes in the state $s_t$ at step $t$ and outputs a boolean value to tell whether $s_t$ is the corresponding key state $\kappa_i$.
This mechanism systematically annotates each state in trajectories as a key state instance or not.
For example, $s_t \mapsto s_t^{\kappa_2}$ in Figure~1(b) denotes that $s_t$ is labeled as an instance of key state $\kappa_2$.

\textbf{Prompt design} aims to (i) alleviate the burden of labor-intensive prompt engineering and (ii) facilitate multi-agent cooperation through labor division.
To do so, we structure the prompt with a standardized prompt template and task information, as shown in Figure~1(b).
Our designed \textit{prompt template} is consistent across tasks and primarily includes role instructions to help LLM understand its role, as well as JSON output constraints.
To encourage \textit{multi-agent cooperation}, we prompt LLM to promote labor division among agents via assigning different roles to agents in key states.
For a symbolic state task, the prompt only requires essential details, i.e., the task description and the state form, which can be easily extracted from the task document without extra processing, making it less demanding than previous methods~\cite{ma2023eureka, du2023guiding}.
LLM then generates key state definitions and discriminator functions, which can be easily extracted from the JSON response.
We also scale it to vision-based tasks in \ref{appsec:imgbased}.

\textbf{A Self-Check mechanism} is adopted to enable LLM's autonomous evaluation and response improvement, mitigating issues such as hallucinated outputs and non-executable code. This design is inspired by recent approaches~\cite{shinn2023reflexion, dhuliawala2023chain, qu2024latent}.
The mechanism comprises two checking operations: (i) LLM rethinking and (ii) code verification. 
The former prompts LLM with a set of queries for self-assessment, ensuring compliance with specified criteria.
The latter verifies the executability of discriminator functions with actual state inputs, providing feedback until all functions are executable.
Notably, LEMAE can be easily extended to incorporate on-training LLM feedback, such as verifying the correctness of identified key states.
Yet, we aim to minimize the computational overhead of integrating LLMs, and current results in Table~\ref{fig:ablate}(c) show that this simple pre-training design already achieves strong robustness and efficiency, leaving further extensions to future work.

\textbf{LEMAE significantly reduces application constraints and costs}, enhancing its practicality in multi-agent RL:
(i) It incorporates task-relevant information into symbolic states while circumventing predefined components such as observation captioners~\cite{du2023guiding} or environment codes~\cite{xie2023text2reward}, which require manual fine-tuning, probably unavailable in practice, or could introduce risk from extra information.
(ii) The reusability of discriminator functions avoids frequent calls, and our method empirically requires fewer than three LLM inferences for a specific task, contributing only a few additional seconds of runtime, which is negligible relative to the hours needed for RL training.

We have adopted typical LLMs, including GPT-4-turbo~\cite{achiam2023gpt} and the open-source Llama-3.3-70B-Instruct~\cite{grattafiori2024llama}, indicating strong adaptability.
Prompt details are in \ref{appsec:prompt}.

\begin{algorithm}[H]
\small
\caption{LEMAE}
\label{alg:LEMAE}
\begin{algorithmic}[1]
    \INPUT LLM $\mathcal{M}$, prompt $\mathcal{P}$, rethinking prompt $\mathcal{P}^{re}$, scaling factors $\alpha, \beta$, randomness $\epsilon_l, \epsilon_h~(\epsilon_l < \epsilon_h)$, max episodes $\mathcal{N}^{max}$, key state number $\mathcal{K}$
    \OUTPUT Policy network $\bm{\pi}_\theta$
    \STATE Initialize policy $\bm{\pi}_\theta$, key states memory tree $\mathcal{T} \Leftarrow [root]$, trajectory replay buffer $\mathcal{D}$ and key states buffer $\mathcal{D}^{ks}$
    \STATE Obtain initial discriminator functions $\{\hat{\mathcal{F}}_i\}_{i=1}^{\mathcal{K}} \Leftarrow \mathcal{M}(\mathcal{P})$
\STATE // Self-Check Mechanism
    \STATE Refine them via self-check
$\{\mathcal{F}_i\}_{i=1}^{\mathcal{K}} \!\Leftarrow\! \mathcal{M}(\mathcal{P}, \{\hat{\mathcal{F}}_i\}, \mathcal{P}^{re}, \text{execute\_error})$
    \FOR{$episode = 1$ to $\mathcal{N}^{max}$}
        \STATE // Explore with Key States Memory Tree (Algorithm~\ref{alg:ksmt})
        \STATE $(\kappa\_chain, \mathcal{T}, \tau) \Leftarrow$ KSMT-Exp($\bm{\pi}_\theta$, $\mathcal{T}$, $\{\mathcal{F}_i\}$, $\epsilon_l$, $\epsilon_h$)
        \STATE $\mathcal{D} \Leftarrow \mathcal{D} \cup \{\tau\}$; $\mathcal{D}^{ks} \Leftarrow \mathcal{D}^{ks} \cup \{\kappa\_chain\}$
        \STATE Sample batch $B = \{\tau_i\}$ from $\mathcal{D}$ and $B^{ks} = \{\kappa\_chain_i\}$ from $\mathcal{D}^{ks}$
        \FOR{each $(\tau, \kappa\_chain)$ in $(B, B^{ks})$}
        \STATE Assign Subspace-based Hindsight Intrinsic Reward along $\kappa\_chain$ sub-trajectories based on Equation~\ref{eqn:hir}
            \STATE // Plan with Key States Memory Tree
            \STATE Sample a planning node $\xi_i$ from children of branch $\kappa\_chain$ in $\mathcal{T}$ if it exists, otherwise from nodes not in $\kappa\_chain$
            \STATE Let $\kappa_{plan}$ denote the key state corresponding to $\xi_i$
            \STATE Propagate intrinsic rewards toward $\kappa_{plan}$ for the remaining trajectory:
            \[
                r_t \Leftarrow \alpha r_t + \beta r_I^{\kappa_{plan}}(s_t, s_{t+1})
            \]
            \STATE // MARL Training (Algorithm-agnostic)
            \STATE Train $\theta$ with $B$ using an MARL algorithm
        \ENDFOR
    \ENDFOR

\end{algorithmic}
\end{algorithm}

\subsection{Key State-Guided Exploration}\label{sec:exploremethod}

\subsubsection{Subspace-based Hindsight Intrinsic Reward}

In multi-agent systems, reward design often suffers from credit assignment ambiguity and sparse feedback, making it difficult for agents to learn coordinated behaviors.
To alleviate these issues and reduce the burden of manual reward design, key states are used as explicit coordination signals that densify rewards and provide shared policy guidance across agents.
Given the annotated key states, global trajectories can be naturally segmented into sub-trajectories associated with distinct cooperative phases.
Inspired by HER~\cite{andrychowicz2017hindsight}, 
each key state is treated as a subgoal for its corresponding sub-trajectory, enabling the generation of hindsight intrinsic rewards that encourage agents to progress toward meaningful intermediate coordination outcomes, as further detailed in \ref{appsec:her}.

In addition, the state vector indices used by the discriminator reveal a reward-relevant subspace of the full state.
Accordingly, we define the Subspace-based Hindsight Intrinsic Reward (SHIR) as:
\begin{equation}
r_I^{\kappa_m}(t) = \Vert \Phi_m(s_t) - \Phi_m(\kappa_m)\Vert-\Vert \Phi_m(s_{t+1}) - \Phi_m(\kappa_m)\Vert,
\label{eqn:hir}
\end{equation}
where $\Vert\cdot\Vert$ denotes a distance metric, e.g., Manhattan distance; $\Phi_m(s)=(s_e)_{e\in\upsilon_m}$ restricts the state space to elements $e\in\upsilon_m$, $s_e$ is the $e$-th element of the full-state $s$, and $\upsilon_m\subset\mathbb{N}^+$  refers to the reward-relevant subset of state space used in the discriminator function $\mathcal{F}_m$.

\textbf{Potential per-agent allocation.} 
In principle, the subspace decomposition provided by the LLM can be further mapped to individual agents, enabling per-agent intrinsic rewards:
\begin{equation}
r_{I,i}^{\kappa_m}(t) = \Vert \Phi_{m,i}(s_t) - \Phi_{m,i}(\kappa_m)\Vert - \Vert \Phi_{m,i}(s_{t+1}) - \Phi_{m,i}(\kappa_m)\Vert,
\end{equation}
where $\Phi_{m,i}(s)$ restricts the subspace $\upsilon_m$ to the dimensions controlled or observed by agent $i$. This formulation would allow fine-grained, agent-specific guidance toward coordination-critical states.

However, in multi-agent settings, assigning intrinsic rewards individually reduces the problem to independent learning, which often suffers from high variance and non-stationarity.
By contrast, combining SHIR with a value decomposition~\cite{rashid2018qmix} framework naturally aggregates credit assignment across agents, providing stable and consistent learning signals. 
As a result, our current implementation shares intrinsic rewards across agents, i.e., Equation~\ref{eqn:hir}, which empirically achieves robust and efficient learning, while retaining the theoretical flexibility for future per-agent extensions.

\textbf{Adopting subspace-based rewards} reduces redundancy and mitigates the bias induced by designing intrinsic rewards over the entire state space, since cooperative rewards in MARL often depend on limited, coordination-critical dimensions~\cite{liu2021cooperative, todorov2012mujoco}.
LEMAE is also applicable to scenarios where rewards depend on the entire state space, as it imposes no strict constraints.
For time step $t$, let $m_t$ denote the index of the key state 
terminating the sub-trajectory to which $t$ belongs. 
The final reward function is given by:
\begin{equation}
    r(t) = \alpha\cdot r_E(t) + \beta\cdot  r_I^{\kappa_{m_t}}(t),
\end{equation}
where $r_E$ denotes the extrinsic reward with $\alpha, \beta\in \mathbb{R^+}$ non-negative scaling factors.

\subsubsection{Key States Memory Tree}

To organize multi-agent exploration with memory, we introduce the concept of Key States Memory Tree (KSMT), which tracks transitions between key states.
Beyond simple $\epsilon$-greedy strategies, KSMT enhances exploration efficiency by (i) focusing on meaningful state subspaces and (ii) incorporating subgoal planning to guide cooperation behavior.
This reduces redundant search in the exponentially large MARL state space.
LEMAE is compatible with alternative memory structures, such as Directed Acyclic Graphs.

\textbf{Construct KSMT.}
KSMT is initialized at the root node and expanded by iteratively incorporating key state chains extracted from annotated trajectories, as outlined in Algorithm~\ref{alg:ksmt}.
This process continues until the success state is discovered or all key state transitions are sufficiently captured.

\begin{figure}
  \centering
  \includegraphics[width=0.4\linewidth]{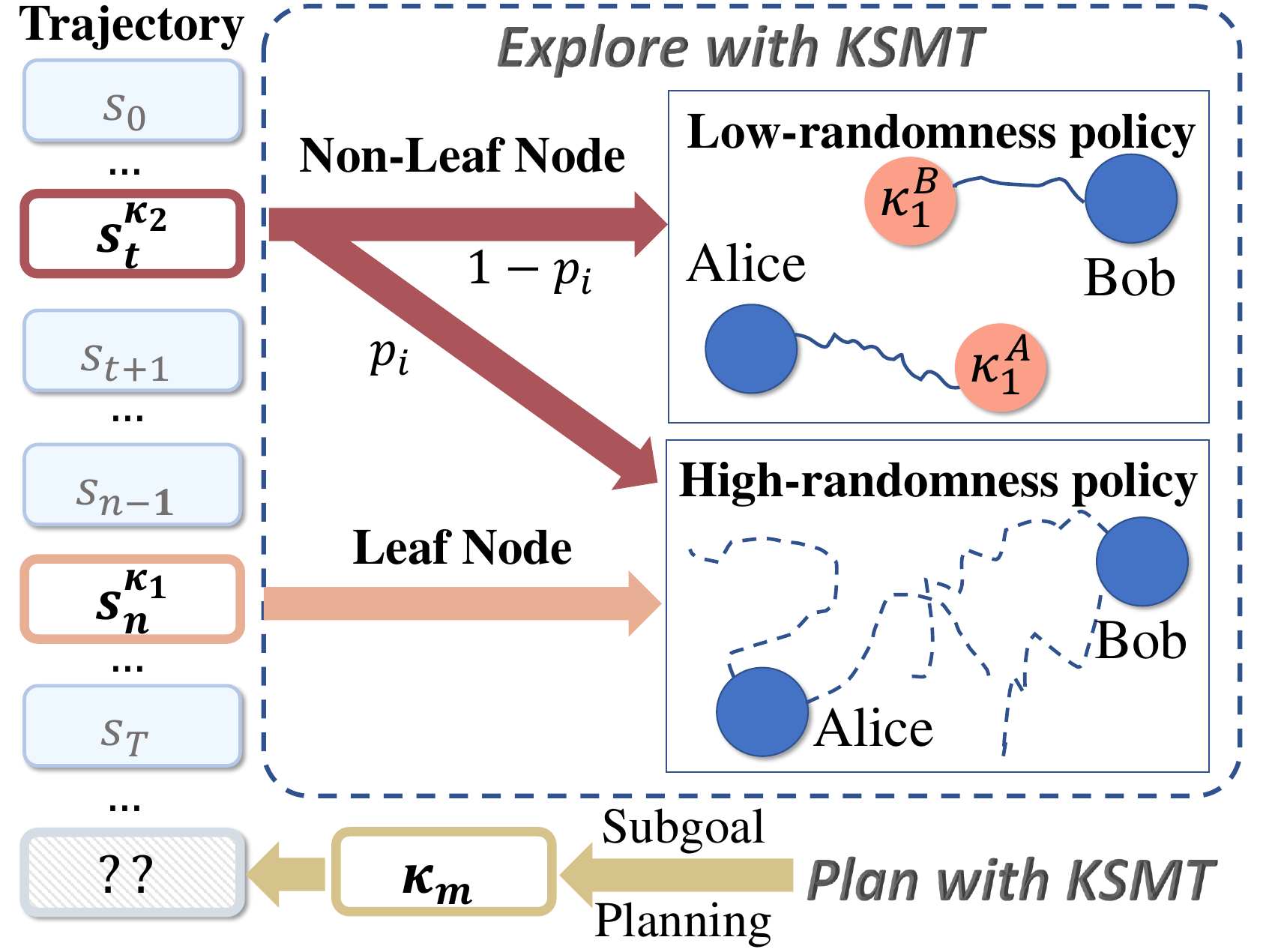}
  \caption{Illustration of KSMT along a trajectory, showing the mixed-randomness strategy and subgoal planning.}
  \label{fig:KSMTE}
\end{figure}
\textbf{Explore with KSMT:}
To expand KSMT, a \textit{mixed-randomness exploration strategy} is devised to balance stochasticity and policy progression, as shown in Algorithm~\ref{alg:ksmt}.
When agents reach a leaf node (a node without children), they adopt a high-randomness policy $\bm{\pi}_\theta^{\epsilon_h}$ to explore new branches.
At a non-leaf node $\xi_i$, agents select $\bm{\pi}_\theta^{\epsilon_h}$ with probability $p_i = \frac{1}{d_i+1}$ or a low-randomness policy $\bm{\pi}_\theta^{\epsilon_l}$ with probability $1-p_i$ to favor steady progression toward the next key state.
$d_i$ denotes the number of outgoing edges (degree) at node $\xi_i$, serving as an indicator of under-exploration.
The exploration phase terminates once a success state is discovered.
Additionally, we \textit{prune branches} that do not lead to success, preventing agents from being distracted by task-irrelevant key states.

\textbf{Plan with KSMT:}
As KSMT serves as a dynamic model within the key state space, it naturally supports \textit{subgoal planning}.
As illustrated in Figure~1(b), given the currently achieved key state chain (e.g., $\kappa_2 \rightarrow \kappa_1$), we perform a lookup operation to locate the corresponding branch in KSMT and its children nodes (e.g., $\kappa_m, ...$).
These children, validated by memory, represent the most probable next key states, from which the final subgoal is randomly sampled.
If no validated child node exist, a key state is instead sampled uniformly from KSMT to ensure the availability of a subgoal.
This mechanism addresses failure cases where trajectories do not reach a key state as the final subgoal.
By doing so, it further enhances the guidance in SHIR and promotes the efficacy of exploring KSMT.

\begin{algorithm}[H]
\footnotesize
\caption{Explore with Key States Memory Tree (KSMT-Exp)}
\label{alg:ksmt}
\begin{algorithmic}[1]
    \INPUT Policy network $\bm{\pi}_\theta$, key states memory tree $\mathcal{T}$, discriminator functions $\{\mathcal{F}_i\}_{i=1}^{\mathcal{K}}$, randomness $\epsilon_l, \epsilon_h~(\epsilon_l < \epsilon_h)$
    \OUTPUT $\kappa\_chain$, updated memory tree $\mathcal{T}$, trajectory $\tau$
    \STATE Initialize $\kappa\_chain \Leftarrow [\ ],\ \tau \Leftarrow \{\}$
    \FOR{$t = 1$ to $t_{\max}$}
        \STATE Discriminate $s_t$ with $\{\mathcal{F}_i\}_{i=1}^{\mathcal{K}}$
        \IF{$s_t$ is a key state $\kappa_m$ \AND $\kappa_m \notin \kappa\_chain$}
            \STATE Append $\{t, \kappa_m\}$ to $\kappa\_chain$
            \IF{branch corresponding to $\kappa\_chain$ not in $\mathcal{T}$}
                \STATE Add the branch into $\mathcal{T}$ \RightComment{update KSMT}
            \ELSIF{$\kappa_m$ corresponds to a non-leaf node $\xi$}
                \STATE $d \Leftarrow$ degree of node $\xi$; $p \Leftarrow 1 / (d + 1)$
                ;$\epsilon \Leftarrow 
                \begin{cases}
                    \epsilon_h & \text{with prob. } p \\
                    \epsilon_l & \text{with prob. } 1 - p
                \end{cases}$
            \ELSE
                \STATE $\epsilon \Leftarrow \epsilon_h$
            \ENDIF
        \ENDIF
        \STATE With probability $\epsilon$, select a random action $\bm{a}_t$; otherwise sample $\bm{a}_t \sim \bm{\pi}_\theta(s_t)$
        \STATE Execute $\bm{a}_t$ to obtain $(s_t, \bm{a}_t, s_{t+1}, r_t)$
        \STATE $\tau \Leftarrow \tau \cup \{(s_t, \bm{a}_t, s_{t+1}, r_t)\}$
    \ENDFOR
\end{algorithmic}
\end{algorithm}

\section{Experiments}

We conduct main experiments on commonly used multi-agent exploration benchmarks: (1) the Multiple-Particle Environment (MPE)~\cite{lowe2017multi, wang2019influence} and (2) the StarCraft Multi-Agent Challenge (SMAC) v1~\cite{samvelyan2019starcraft} and v2~\cite{ellis2024smacv2}, as illustrated in Figure~1(3).
Following previous studies~\cite{ma2023eureka,liu2021cooperative}, 
we focus primarily on tasks with \textbf{symbolic state spaces} and use the \textbf{sparse reward} version for all tasks without specific instructions. 
\textbf{Baselines.}\label{sec:experimentalsetup}
We compare LEMAE with representative methods.
(1) Traditional \textit{\textbf{methods for efficient multi-agent exploration}} without LLMs include EITI, EDTI~\cite{wang2019influence}, 
CMAE~\cite{liu2021cooperative}, MASER~\cite{jeon2022maser},
LAIES~\cite{liu2023lazy},
FOX~\cite{jo2024fox},
ICES~\cite{li2024ices}.
(2) \textit{\textbf{LLM-based methods}}, such as ELLM~\cite{du2023guiding}, which generates exploration goals using LLMs; Eureka~\cite{ma2023eureka}, which uses LLM for reward design; and ProgressCount (ProCnt)~\cite{sarukkai2024automated}, which combines LLM reward design with count-based exploration.
We retain the prompt information consistent across all relevant LLM-based methods.
LEMAE is implemented on IPPO~\cite{de2020independent} in MPE and QMIX~\cite{rashid2018qmix} in SMAC, following previous works~\cite{wang2019influence, liu2023lazy} to ensure fair comparisons.

We run each algorithm on five random seeds and report the mean performance with standard deviation. 
Further details can be referenced in \ref{appsec:expdetails}.

\vspace{-1pt}
\subsection{MPE}\label{exp:mpe}
\vspace{-1pt}

We adopt four commonly used tasks~\cite{wang2019influence, liu2021cooperative}: \textit{Pass}, \textit{Secret-Room}, \textit{Push-Box}, and \textit{Large-Pass}.

\textbf{LLM can discriminate key states.}
We examine the efficacy of LLM in discriminating key states.
On the \textit{Pass} task,
as shown in Figure~1(a), a wall divides the room, with each half containing an invisible switch. 
The door opens only when one agent stands on a switch. Agents must cooperate to move from the left half-room to the right one. 
In Figure~\ref{fig:roomanalyze}(a), LLM exhibits a precise understanding of the task and generates meaningful discriminator functions, validating LEMAE with current LLM capabilities.

\begin{figure}[t]
    \centering
    \vspace{-5pt}
    \includegraphics[width=0.9\linewidth]{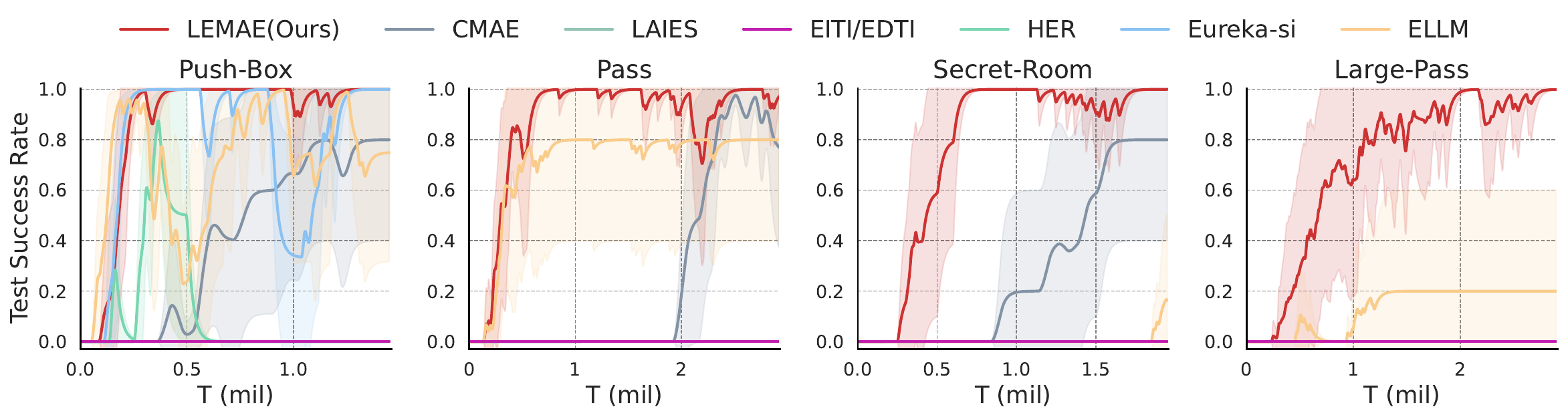}
    \vspace{-10pt}
    \caption{Evaluating LEMAE against baseline methods on four MPE maps with \textbf{sparse rewards}, using test success rate as the evaluation metric.
    }
    \vspace{-6pt}
    \label{fig:roomresult}
\end{figure}

\begin{figure}[t]
\centering
\begin{minipage}[b]{0.24\linewidth}
    \centering
    \includegraphics[width=\linewidth]{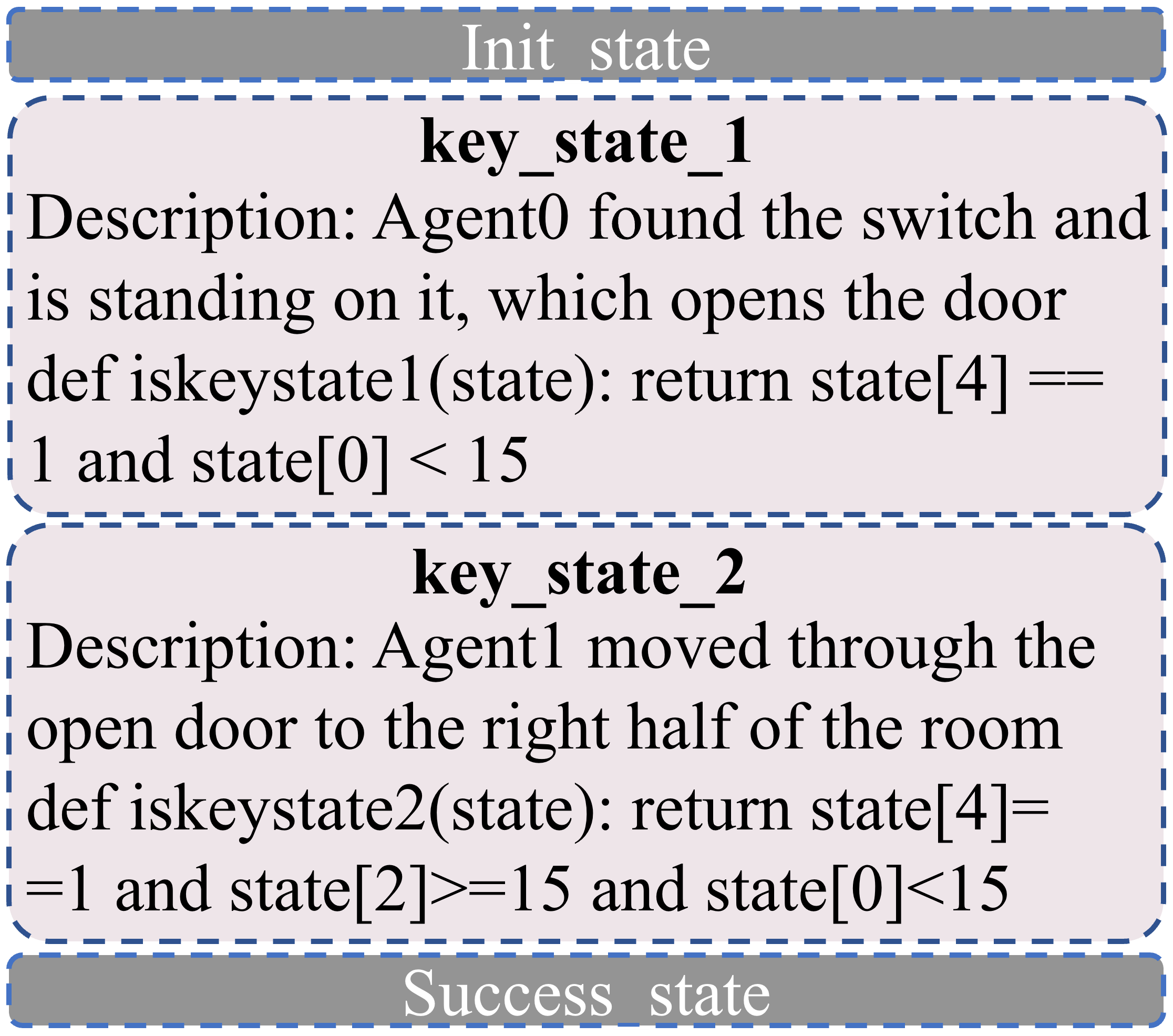}
    {\footnotesize (a)}
    \label{fig:passresponse}
\end{minipage}
\hspace{20pt}
\begin{minipage}[b]{0.21\linewidth}
    \centering
    \includegraphics[width=\linewidth]{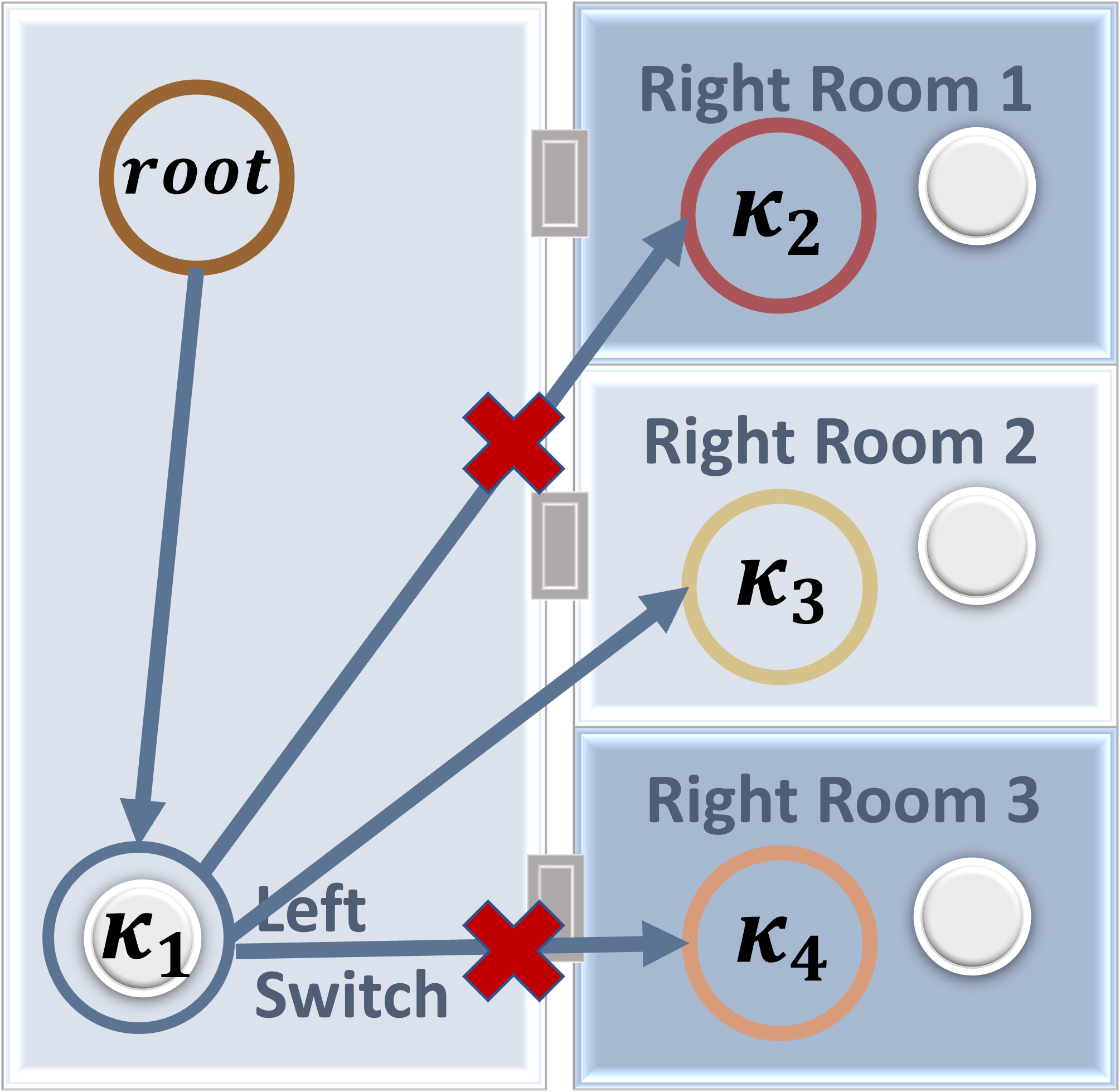}
    {\footnotesize (b)}
    \label{fig:keystatenum-b}
\end{minipage}
\hspace{20pt}
\begin{minipage}[b]{0.24\linewidth}
    \centering
    \includegraphics[width=\linewidth]{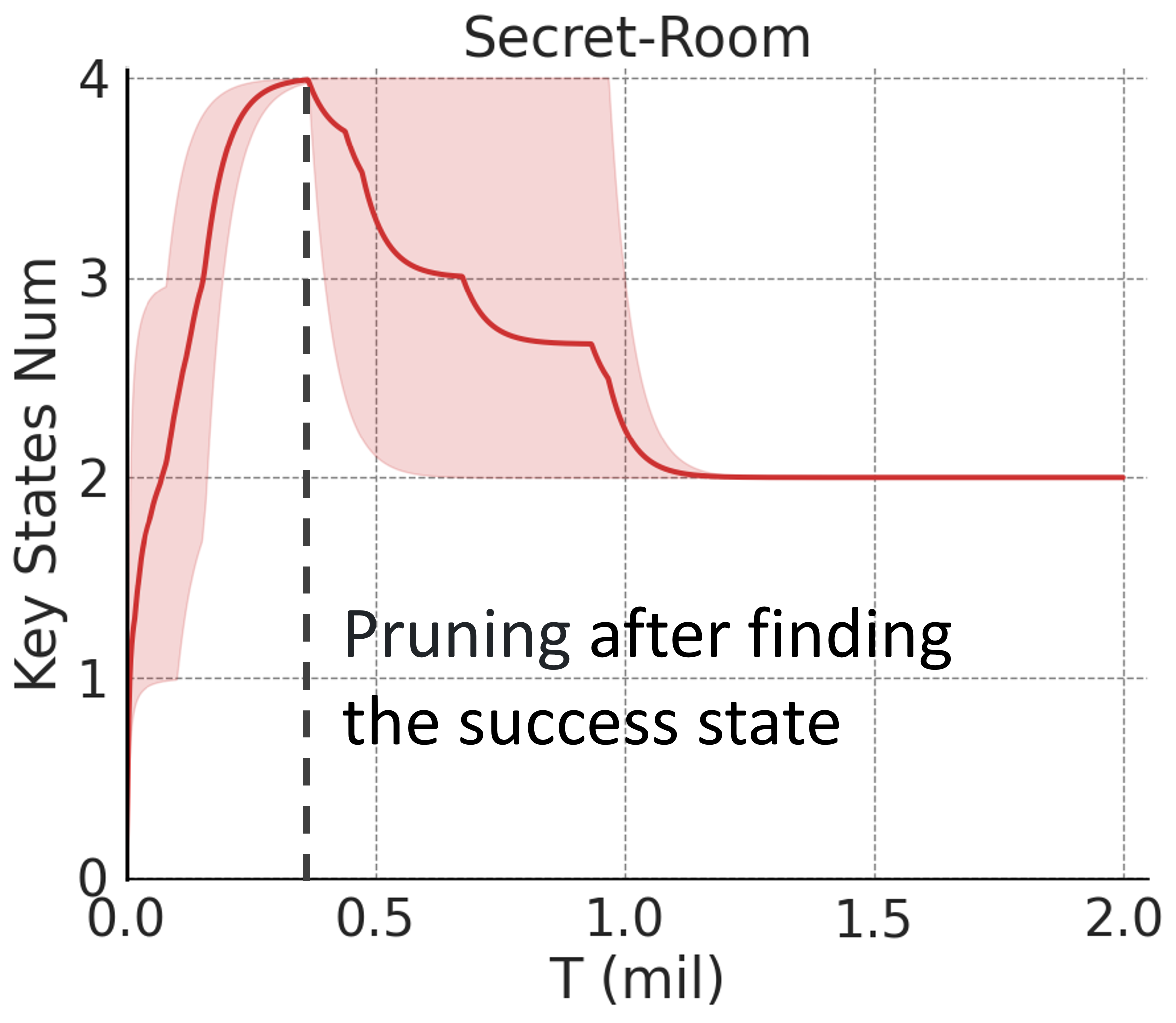}
    {\footnotesize (c)}
    \label{fig:keystatenum-c}
\end{minipage}

\vspace{-6pt}
\caption{
(a) Key states discrimination functions generated on \textit{Pass}. 
(b) \textit{Secret-Room} with key states: \textbf{$\kappa_1$} denotes occupying the left switch to open all doors, while \textbf{$\kappa_{2,3,4}$} represent exploring right rooms 1, 2, and 3, respectively. Arrows symbolize the transition structure within KSMT. 
(c) Key states number curve on \textit{Secret-Room} shows that LEMAE can identify key states and proficiently prune task-irrelevant ones. 
}
\label{fig:roomanalyze}
\vspace{-7pt}
\end{figure}

\textbf{LEMAE achieves superior performance.}
As shown in Figure~\ref{fig:roomresult}, LEMAE outperforms baselines by accelerating exploration through the use of LLM priors.
The failure of common baselines highlights the need for efficient exploration, while LEMAE's superior performance demonstrates the effectiveness of incorporating task-specific guidance from LLMs.
While CMAE learns effective strategies for simple tasks, its redundant exploration hinders efficiency, making it inadequate for tasks with large exploration spaces such as \textit{Large-Pass}.
Although ELLM benefits from LLM priors, it lags behind LEMAE due to weak reward guidance, not to mention its reliance on frequent LLM calls and predefined state captioners.
Furthermore, we compare LEMAE with CMAE using the metric of the number of exploration steps to reach a success state.
Results show a \textbf{substantial acceleration—up to 10×}—highlighting LEMAE’s efficiency.
This superior performance stems from its ability to reduce redundant exploration by integrating task-relevant priors.

\begin{table}[htbp]
  \centering
  \caption{Comparison of LEMAE and CMAE, the traditional SOTA MAE method, in terms of exploration steps (in thousands) needed to find the success state.}
  \resizebox{0.5\linewidth}{!}{
    \begin{tabular}{cccc}
    \toprule
    Tasks & LEMAE~(Ours) & CMAE & Acceleration rate \\
    \midrule
    Push-Box & \textbf{159.0\scriptsize{$\pm$42.5}} & 972.3\scriptsize{$\pm$887.3} & \textbf{6.1$\times$} \\
    Pass & \textbf{153.1\scriptsize{$\pm$20.7}} & 2114.8\scriptsize{$\pm$157.4} & \textbf{13.8$\times$} \\
    Secret-Room & \textbf{316.6\scriptsize{$\pm$134.6}} & 1448.5\scriptsize{$\pm$467.2} & \textbf{4.6$\times$} \\
    Large-Pass & \textbf{446.9\scriptsize{$\pm$256}} & \textgreater3000 & \textbf{\textgreater6.7$\times$} \\
    \bottomrule
    \end{tabular}
    }%
    \vspace{-5pt}
  \label{tab:firstsuccess}%
\end{table}%

\textbf{LEMAE benefits from LLM priors through discrimination.}
We evaluate HER~\cite{andrychowicz2017hindsight}, which proposes hindsight intrinsic rewards but selects goals randomly from memory.
Its poor performance underscores the importance of incorporating LLM priors for efficient exploration.
To support our claim about the superiority of LLM discrimination over generation, we evaluate Eureka-si, a single-iteration Eureka variant, which uses LLM to generate reward functions. 
While Eureka-si performs well in simple tasks, it struggles in complex tasks with partial observability, indicating that LLM-based discrimination offer a more general and effective integration of LLM. 
Notably, neither method is specifically designed for efficient exploration.
Please refer to \ref{appsec:her}, \ref{appsec:llmrd} for details.

\textbf{LEMAE reduces redundant exploration.}
We compare the exploration behavior of LEMAE and CMAE on \textit{Pass}. 
The visitation maps in Figure~1(a) reveal that LEMAE markedly reduces redundant exploration: CMAE-trained agents excessively explore the left room, while LEMAE-trained agents' visitation area is concentrated around the success path. 
Furthermore, an organic division of labor among agents emerges, affirming the efficacy of encouraging cooperation in prompt design.

\textbf{LEMAE circumvents task-irrelevant key states.}
Due to incomplete information, LLM may discriminate task-irrelevant key states. 
For instance, in \textit{Secret-Room}, three rooms are present on the right, but LLM is not informed about the real target room for fairness. 
In Figure~\ref{fig:roomanalyze}(b), LLM discriminates two task-irrelevant key states, \textbf{$\kappa_2$} and \textbf{$\kappa_4$}, corresponding to two irrelevant rooms.
Figure~\ref{fig:roomanalyze}(c) illustrates that KSMT's pruning mechanism, activated after finding the success state, effectively circumvents these irrelevant key states.
A more detailed robustness analysis can be found in \ref{sec:robust}.

\begin{figure}[t]
  \centering
    \includegraphics[width=\linewidth]{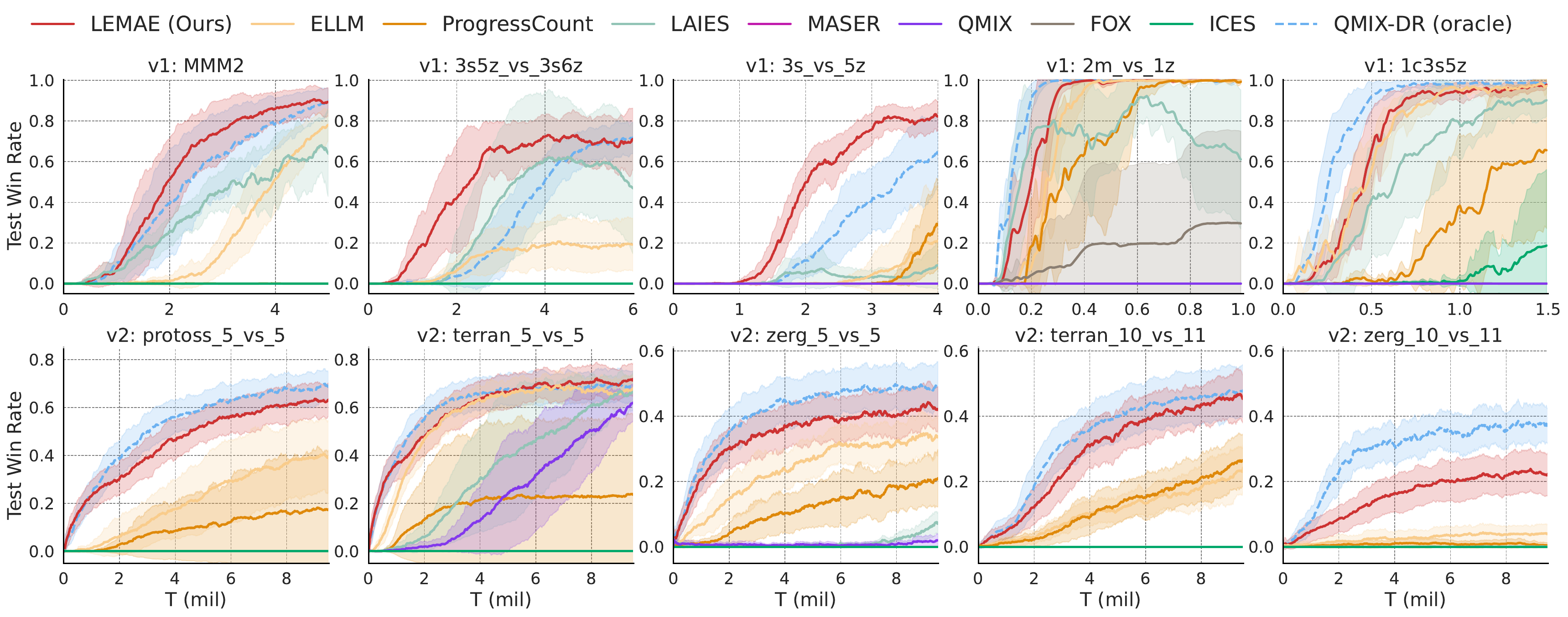}
    \vspace{-18pt}
    \caption{
Evaluating LEMAE on SMAC v1 and v2 with \textbf{sparse rewards}, using test win rate as the evaluation metric. \textbf{QMIX-DR} refers to QMIX trained with dense rewards in the original SMAC environment, serving as an oracle baseline.}
\vspace{-10pt}
    \label{fig:sc2result}
\end{figure}
\subsection{SMAC v1 and v2}\label{exp:smac}
\vspace{-1pt}
SMAC is a widely-used challenging benchmark in MARL. 
Instead of dense or semi-sparse reward versions used before, we employ \textbf{fully sparse-reward tasks} to emphasize exploration, rewarding agents only upon complete enemy elimination.
We validate LEMAE across diverse scenarios with varied difficulty and agent numbers.
SMACv2 is an enhanced version with more stochasticity.

In Figure~\ref{fig:sc2result}, LEMAE demonstrates superior performance.
While baselines like QMIX excel in dense or semi-sparse reward settings, they struggle in fully sparse setups.
CMAE performs well on simple tasks but fails in harder ones due to the lack of task-related information in curiosity-driven goal selection.
LAIES is the best non-LLM baseline, but it relies on handcrafted state priors and still underperforms.
ELLM and ProgressCount perform well on simpler tasks with LLM priors, but their effectiveness diminishes on harder tasks, likely due to the instability of similarity-based rewards and the inefficiency of undirected count-based exploration.
Notably, we add \textbf{QMIX-DR}, which augments QMIX with dense rewards in the original SMAC. 
Surprisingly, LEMAE matches or even surpasses QMIX-DR, highlighting its potential to reduce manual effort in complex reward design.
The superior performance of LEMAE on tasks involving more than ten agents, such as \textit{MMM2} and \textit{10\_vs\_11}, demonstrates its effectiveness in handling relatively large agent populations.
Moreover, given the complexity of the SMAC benchmark, especially the stochastic nature of SMACv2, LEMAE’s consistent superiority confirms its potential applicability in more complex real-world scenarios.

\begin{figure}[t]

\begin{minipage}[b]{0.445\linewidth}
    \centering
    \includegraphics[width=\linewidth]{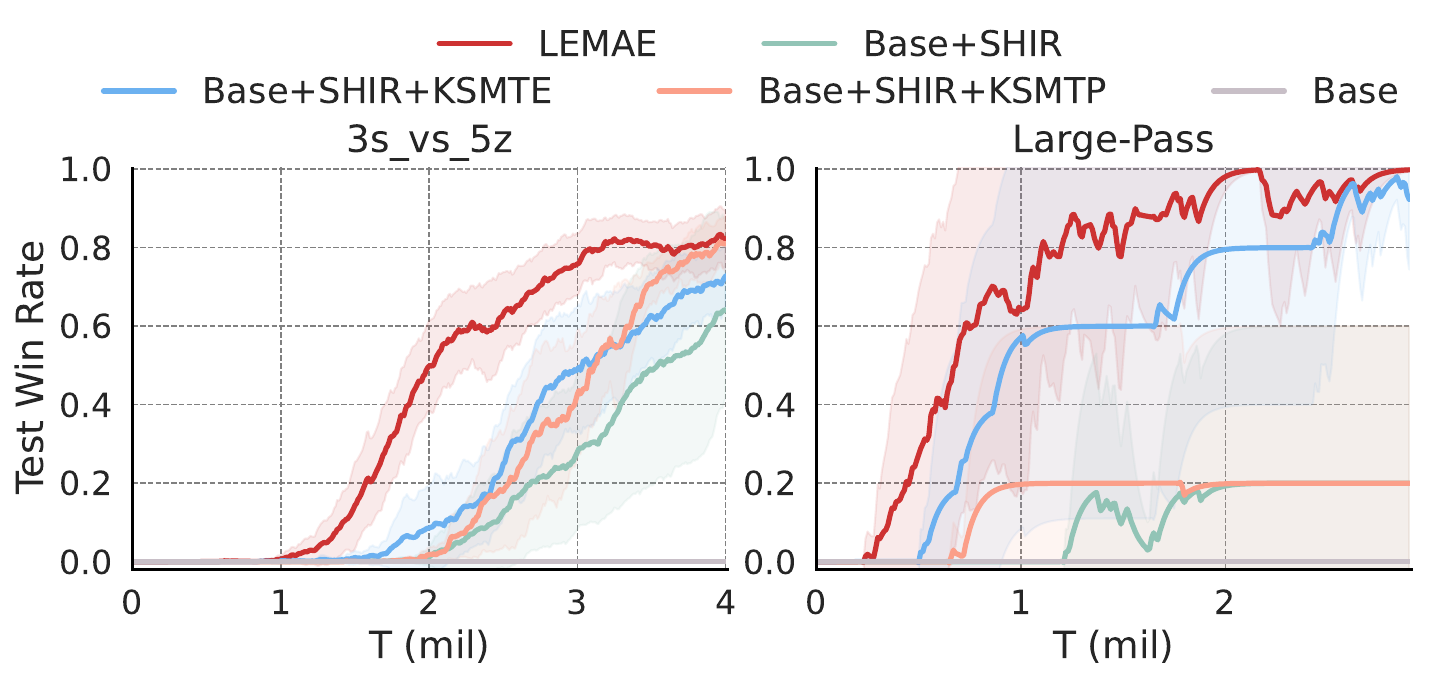}
    \vspace{-12pt}
    \label{fig:sc2ablation}
\end{minipage}
\hspace{0pt}
\begin{minipage}[b]{0.22\linewidth}
    \centering
    \includegraphics[width=\linewidth]{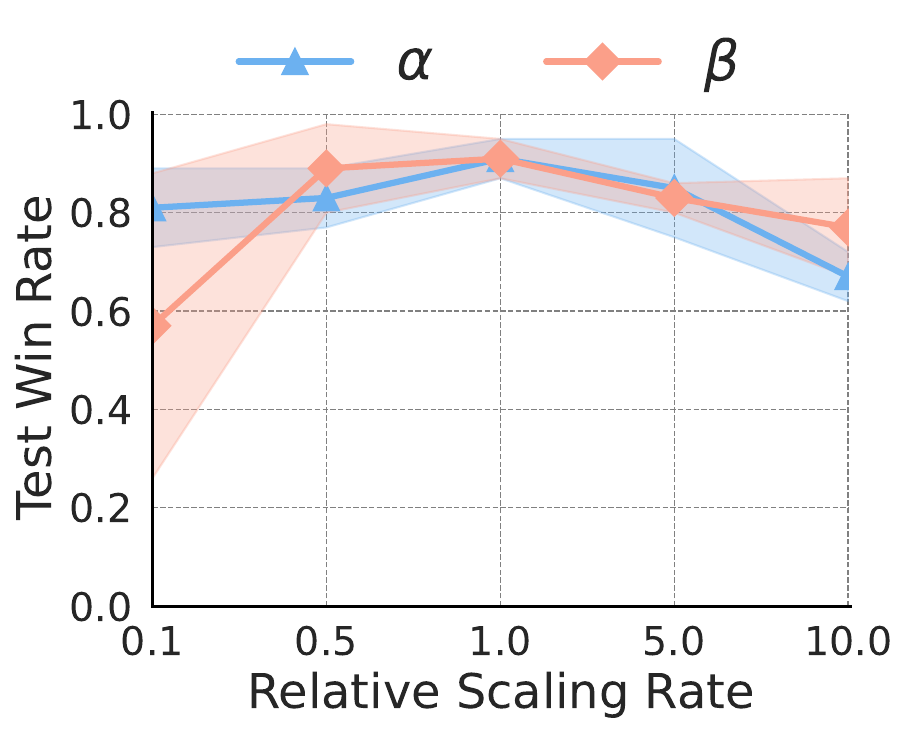}
    \vspace{-8pt}
    \label{fig:hyperablation}
\end{minipage}
\hspace{0pt}
\begin{minipage}[b]{0.31\linewidth}
    \centering
    \resizebox{\linewidth}{!}{
    \begin{tabular}{cccc}
    \toprule
    \large{$r_{acc}$} \small{($r_{exe}$)} & \footnotesize{GPT-4} & \makecell{\footnotesize{GPT-4}\\\footnotesize{w/o}\\\footnotesize{Self-Check}} & \footnotesize{Llama-3.3} \\
    \midrule
    Large-Pass & \textbf{1.0 \scriptsize{(1.0)}} & 0.8 \scriptsize{(1.0)} & \textbf{1.0 \scriptsize{(1.0)}} \\
    2m\_vs\_1z & \textbf{1.0 \scriptsize{(1.0)}} & 0.7 \scriptsize{(1.0)} & \textbf{1.0 \scriptsize{(1.0)}}  \\
    MMM2 & \textbf{0.8 \scriptsize{(1.0)}} & 0.6 \scriptsize{(0.7)} & 0.7 \scriptsize{(1.0)} \\
    \bottomrule
    \end{tabular}
    }
    \vspace{10pt}
    \label{tab:scllmablation}
\end{minipage}

\vspace{-6pt}
\caption{
(a) Ablation studies on \textbf{KSMT and SHIR} on two representative tasks from MPE and SMAC. "Base" refers to the backbone algorithms, and "KSMT-E/P" denotes Explore and Plan with KSMT, respectively.
(b) \textbf{Hyperparameter analysis} of the reward scaling factors $\alpha$ and $\beta$, where the x-axis shows values relative to the default settings.
(c) Ablation studies on the \textbf{Self-Check mechanism and LLM choice}, comparing GPT-4-turbo and Llama-3.3 by reporting the acceptance rate ($r_{acc}$) and execution rate ($r_{exe}$) over ten runs of discriminator generation.
}
\label{fig:ablate}
\vspace{-5pt}
\end{figure}

\begin{figure}[t]

\begin{minipage}[b]{0.29\linewidth}
    \centering
    \includegraphics[width=\linewidth]{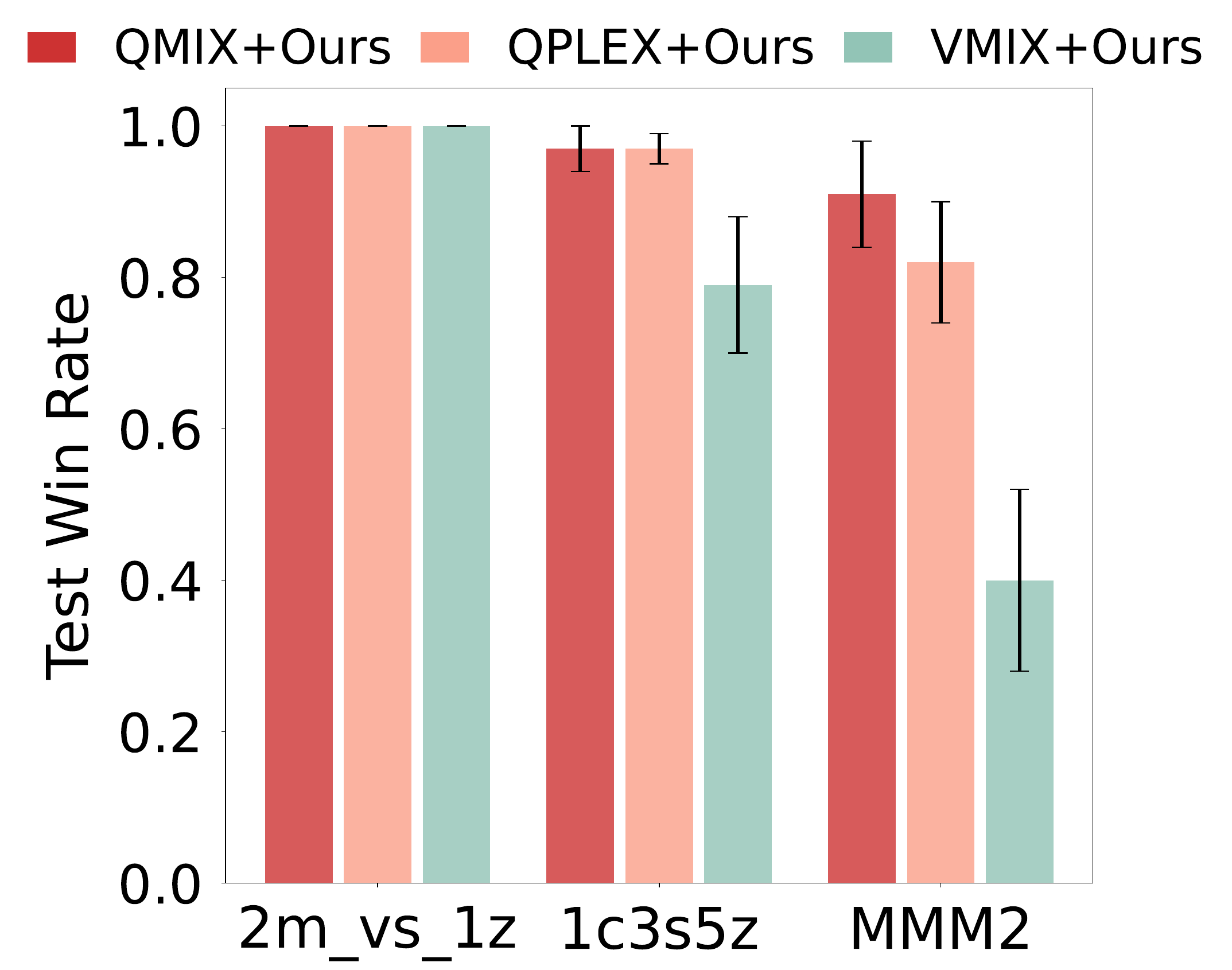}
    \rule{0pt}{13pt} 
    \label{fig:algo}
\end{minipage}
\hfill
\begin{minipage}[b]{0.71\linewidth}
    \centering
    \includegraphics[width=\linewidth]{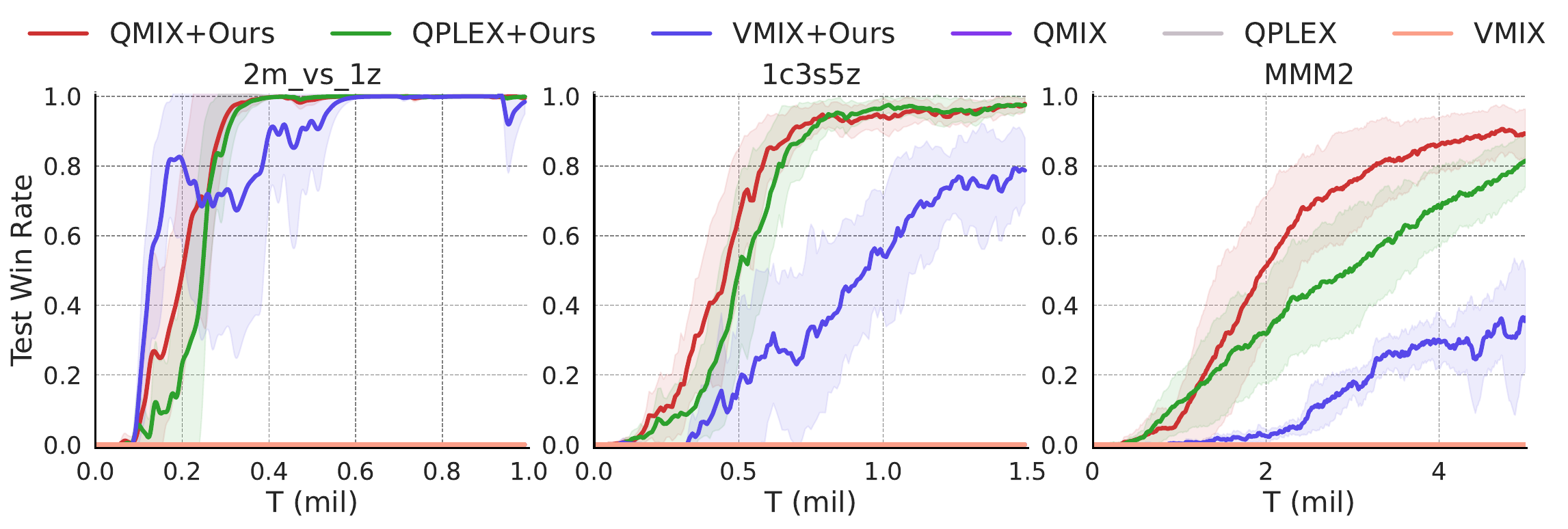}
    \label{fig:algocurve}
\end{minipage}

\vspace{-15pt}
\caption{
LEMAE is compatible with various MARL algorithms, e.g., \textbf{QPLEX} and \textbf{VMIX}. Notably, both QPLEX and VMIX exhibit complete failure unless integrated with our approach.
}
\label{fig:algocombine}
\end{figure}

\subsection{Compatiblility with Various Algorithms}
LEMAE incorporates task-relevant guidance through intrinsic rewards and is agnostic to RL algorithms.
\ref{exp:mpe} and \ref{exp:smac} verified its compatibility with two MARL algorithms: IPPO in MPE and QMIX in SMAC. 
To further substantiate the claim, we build LEMAE on two widely-used MARL algorithms, QPLEX~\cite{wang2020qplex} and VMIX~\cite{su2021value}, representing value-based and actor-critic paradigms, respectively.
As shown in Figure~\ref{fig:algocombine}, algorithms combined with LEMAE consistently outperform baselines, highlighting its adaptability to integrate with other algorithms in diverse applications.
Moreover, LEMAE shows potential as a versatile  exploration mechanism applicable beyond MARL.
We evaluate LEMAE in a \textbf{single-agent variant of MPE}, as detailed in \ref{appsec:single-agent}.

\subsection{Ablation Studies}


\textbf{Role of SHIR and KSMT.}
We conduct an ablation study to evaluate the contributions of SHIR and KSMT. 
We select two exemplary tasks from MPE and SMAC and report results in Figure~\ref{fig:ablate}(a). 
In SMAC, Base refers to QMIX, while in MPE, it denotes IPPO. 
\textbf{KSMTE} signifies exploration with KSMT, \textbf{KSMTP} denotes planning with KSMT, and LEMAE encompasses Base+SHIR+KSMTE+KSMTP.
As shown, removing either SHIR or KSMT leads to a substantial drop in performance, highlighting the essential roles of both components in enabling effective key state-guided exploration.

\textbf{Role of Self-Check mechanism and LLMs.}
We evaluate GPT-4-turbo and the open-source Llama-3.3-70B-Instruct~\cite{grattafiori2024llama} in generating discriminator functions.
Meanwhile, we examine GPT-4-turbo without the Self-Check mechanism.
The Acceptance Rate~($r_{acc}$) denotes the proportion of seeds achieving over $80\%$ of the best RL performance, and the Execution Rate~($r_{exe}$) measures the proportion of seeds where all discriminator functions are executable.
As shown in Table~\ref{fig:ablate}(c), results highlight that combining a powerful LLM with our Self-Check mechanism ensures high-quality key states, reflected in both the code executability and the final performance.
The scalability of LEMAE with LLMs, including open-source models like Llama-3.3, 
demonstrates its scalability and potential for future application in more complex, real-world tasks with stronger LLMs.

\begin{figure}[t]
\centering
\begin{minipage}[b]{0.46\linewidth}
    \centering
    \includegraphics[width=\linewidth]{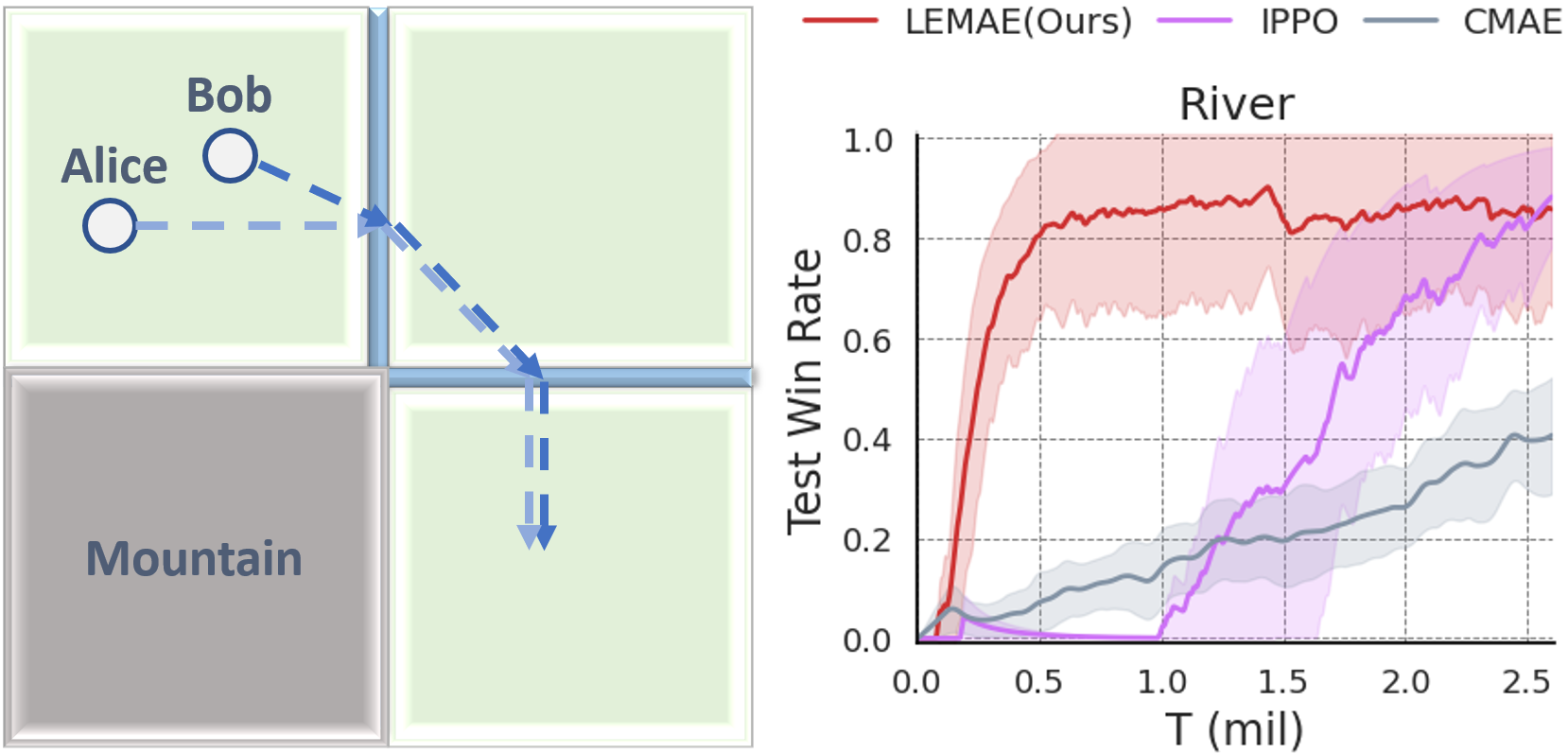}
    {\footnotesize (a)}
    \label{fig:rivercurve}
\end{minipage}
\hspace{15pt}
\begin{minipage}[b]{0.46\linewidth}
    \centering
    \includegraphics[width=\linewidth]{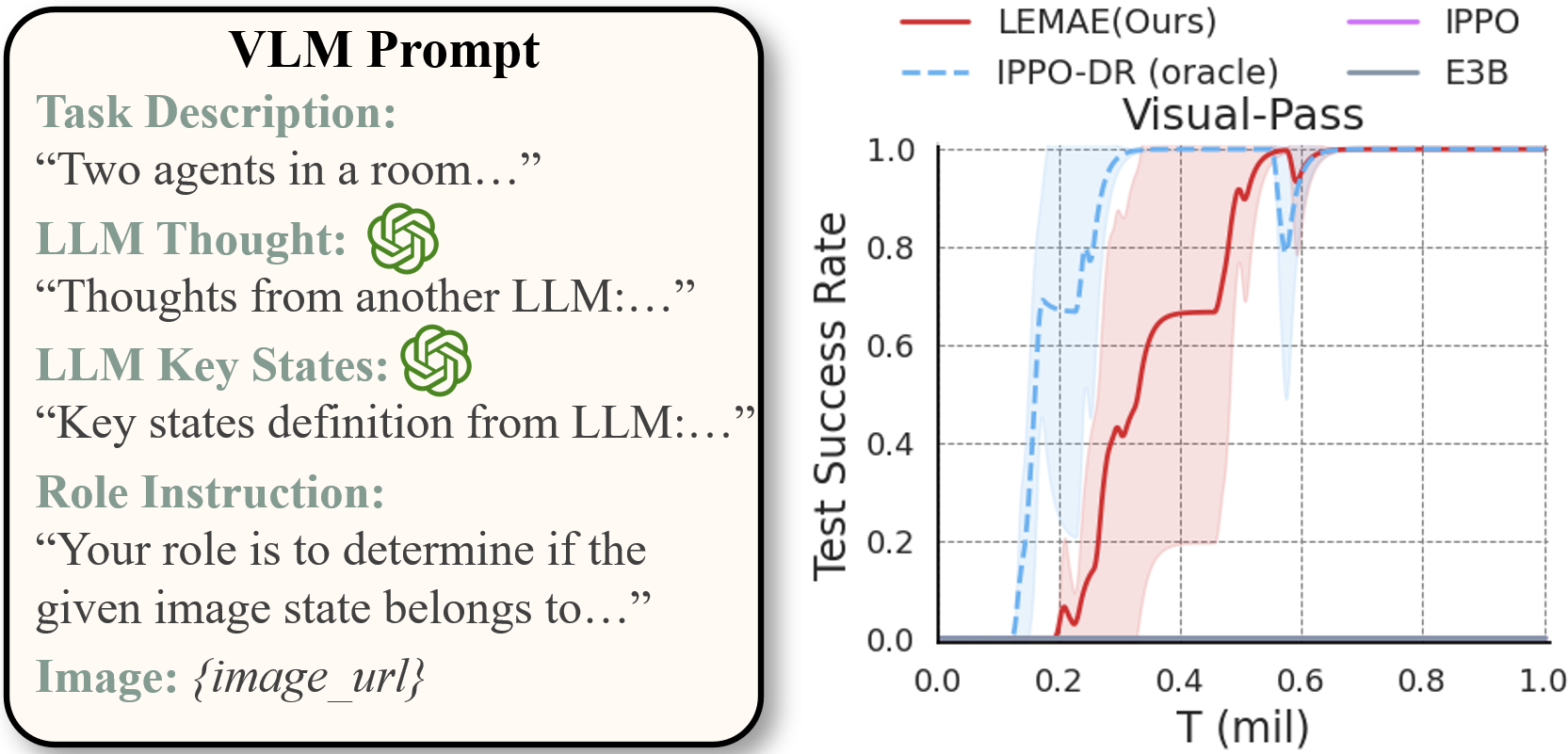}
    {\footnotesize (b)}
    \label{fig:visualcurve}
\end{minipage}

\vspace{-6pt}
\caption{
(a) LEMAE achieves efficient exploration on a \textbf{LLM-unseen task, \textit{River}}. 
(b) LEMAE extends to the \textbf{image-based task \textit{Visual-Pass}}. 
\textbf{IPPO-DR} denotes IPPO trained with human-designed dense rewards, serving as an oracle baseline.
}
\label{fig:extension}
\end{figure}
\subsection{Scalability \& Generalization Analysis}

\textbf{Generalization to Unseen Tasks.}
To ensure LEMAE's success doesn't stem from LLM's prior exposure to the tasks, we evaluate it on a novel, LLM-unseen task, \textit{River} (details in \ref{appsec:river}). 
Figure~\ref{fig:extension}(a) shows LEMAE retains strong performance, confirming that LLM's generalization ability effectively supports LEMAE in facilitating efficient exploration across diverse and novel environments.

\textbf{Extension to Vision-based Tasks.}
While this work centers on symbolic state tasks, LEMAE can be extended to vision-based tasks by replacing the LLM with a multi-modal model.
As detailed in \ref{appsec:imgbased}, we extend the task \textit{Pass} to a vision-based task \textit{Visual-Pass}, where the observation space consists of the environment map rendered with two agents.
To adapt LEMAE to the visual task, as shown in Figure~\ref{fig:extension}(b), we prompt LLM to define key states with the same task description and role instruction as proposed in \ref{sec:LLMprompt} and use the LLM-generated definition as the prompt for a \textbf{Vision Language Model (GPT-4o)}~\cite{hurst2024gpt4o}.
As shown in Figure~\ref{fig:extension}(b), LEMAE achieves superior performance over both the IPPO baseline and the generic intrinsic reward method E3B~\cite{E3B}, and performs comparably to IPPO-DR, an oracle baseline with dense rewards.
These results highlight the broader applicability and generality of LEMAE beyond symbolic domains.

\begin{figure}[t]
\centering
\begin{minipage}[b]{0.41\linewidth}
    \centering
    \includegraphics[width=\linewidth]{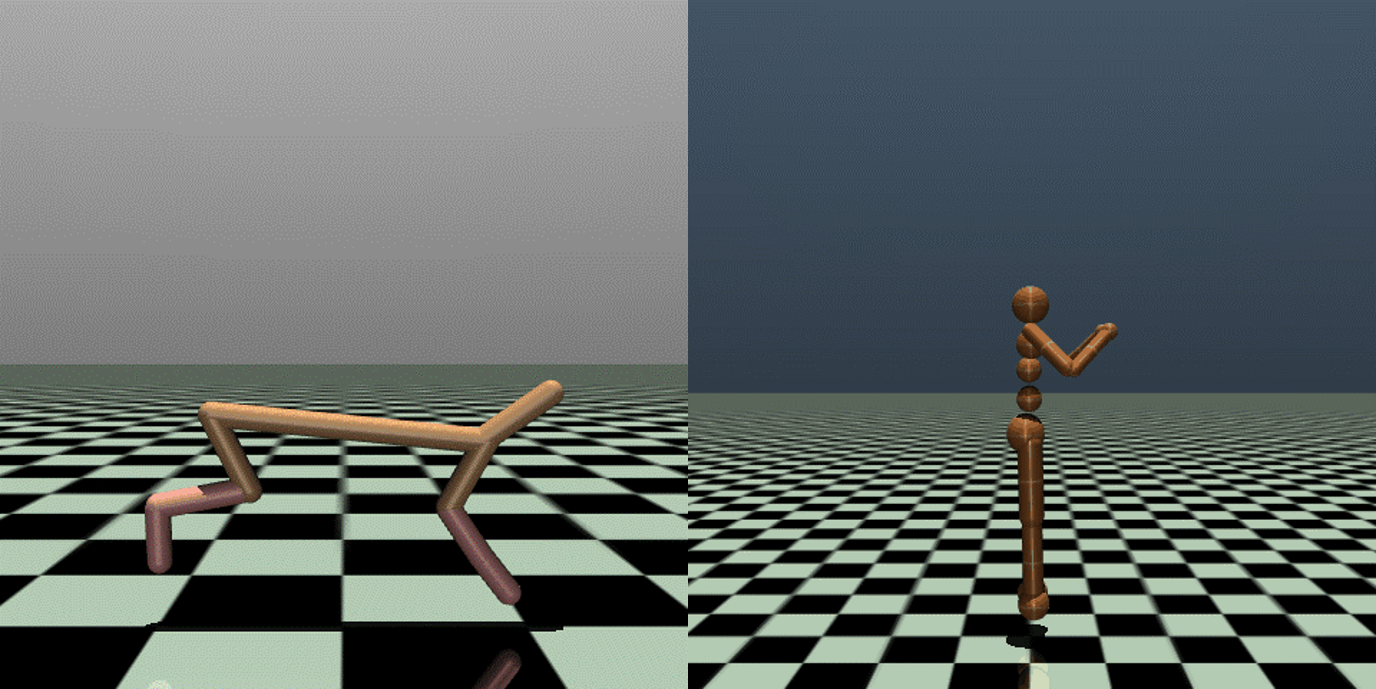}
    \rule{0pt}{25pt}
    \label{fig:envmamujoco}
\end{minipage}
\hspace{0pt}
\begin{minipage}[b]{0.54\linewidth}
    \centering
    \includegraphics[width=\linewidth]{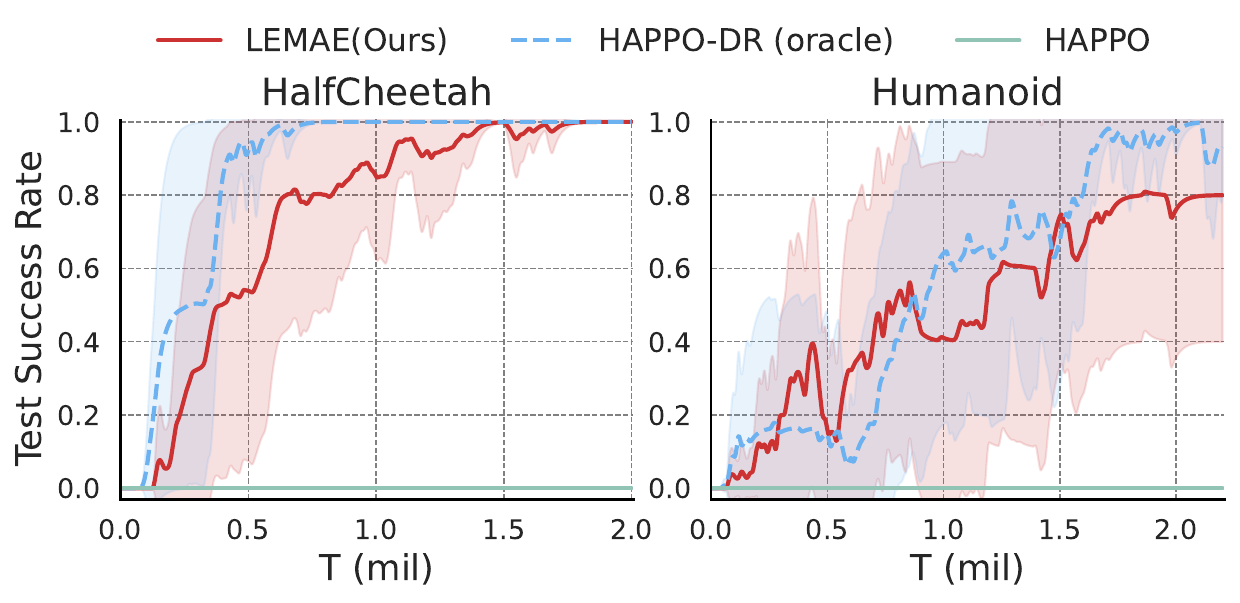}
    \label{fig:mamujococurve}
\end{minipage}

\vspace{-20pt}
\caption{
LEMAE is evaluated on two tasks in \textbf{MAMuJoCo}, a \textbf{MARL robotics} benchmark. 
\textbf{HAPPO-DR} denotes HAPPO trained with human-designed dense rewards, serving as an oracle baseline.
}
\label{fig:mamujoco}
\vspace{-5pt}
\end{figure}

\textbf{Application to Robotic Control.}
We further evaluate LEMAE on two continuous-control tasks, \textit{HalfCheetah} and \textit{Humanoid}, in MaMuJoCo~\cite{peng2021facmac}, a widely used MARL benchmark for robotic locomotion.
In these settings, agents receive only sparse rewards of the form $I(\text{velocity}>\text{threshold})$, where they must coordinate multiple body joints to accelerate and maintain forward motion.
As shown in Figure~\ref{fig:mamujoco}, when integrated with the HAPPO~\cite{kuba2021trust} backbone, LEMAE discovers meaningful velocity subgoals and accelerates training, achieving performance comparable to that of dense-reward-trained baselines.
The results highlight the potential of LEMAE to enhance sample efficiency in complex robotic systems where dense supervision is difficult or costly to obtain.

\begin{figure}[t]
  \centering
  \vspace{-5pt}
    \includegraphics[width=0.65\linewidth]{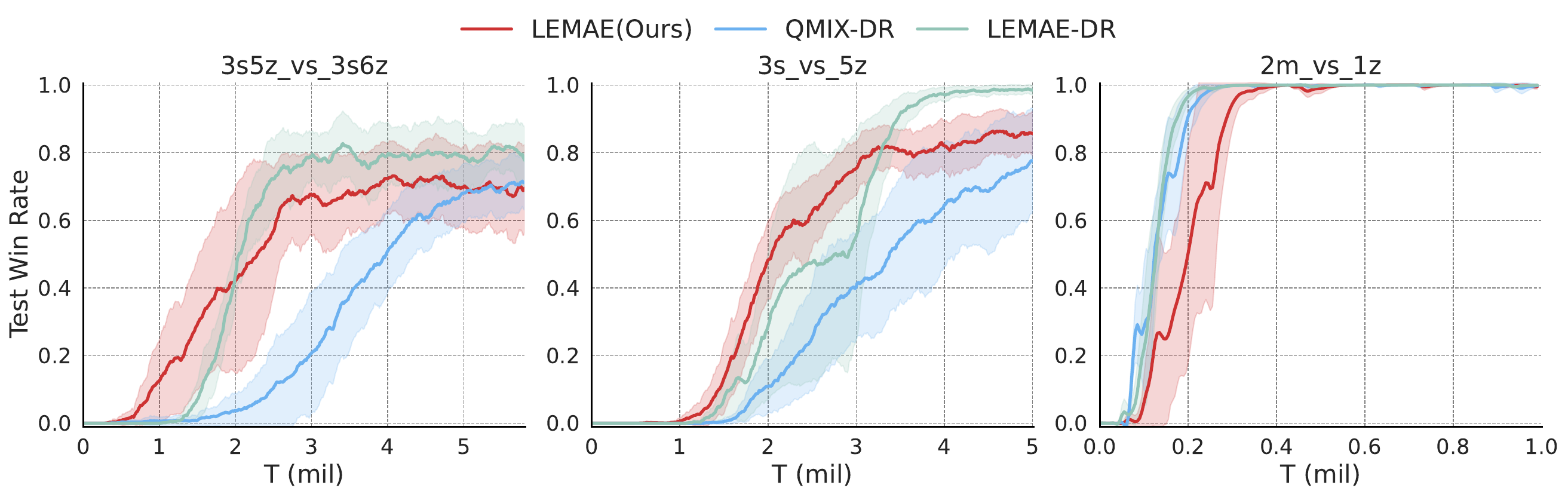}
    \vspace{-5pt}
    \caption{Evaluating LEMAE with \textbf{dense rewards} across three maps in SMAC using the evaluation metric of test win rate. \textbf{LEMAE-DR} is LEMAE with dense rewards in the original SMAC, which effectively ensures efficient exploration and achieves better convergence. }
    \label{appfig:sc2dense}
\end{figure}

\textbf{Working with Dense Reward Settings.}
We also evaluate LEMAE in tasks with dense rewards in SMAC, denoted as LEMAE-DR. As shown in Figure~\ref{appfig:sc2dense}, 
the results confirm that LEMAE-DR facilitates efficient exploration in \textbf{both dense and sparse reward} settings, highlighting the main contribution of our method. Additionally, LEMAE-DR achieves better convergence than LEMAE due to the guidance provided by dense rewards.

\vspace{-3pt}

\subsection{Sensitivity \& Robustness Analysis}\label{sec:robust}
\vspace{-1pt}
\textbf{Sensitivity to Hyperparameters.}
The pivotal hyperparameters in LEMAE are reward scaling rates $\alpha$ and $\beta$.
Figure~\ref{fig:ablate}(b) illustrates that LEMAE is robust to these hyperparameters across a considerable range. 
The x-axis represents the relative values with respect to the default~($\alpha=10$, $\beta=1$), encompassing evaluations for $\alpha\in\{1,5,10,50,100\}$ and $\beta\in\{0.1,0.5,1,5,10\}$. 
Notably, excessive extrinsic reward scaling rate $\alpha$ or insufficient intrinsic reward scaling rate $\beta$ can cause performance degradation due to the abrupt alteration of the reward or the inadequate motivational impact.
Additionally, we conduct an ablation study on mixed-randomness exploration in \ref{appsec:mixrand}.

\textbf{Robustness to Perturbations in Key States.}
We simulate key state perturbations to assess LEMAE's robustness.
Specifically, Reduction mimics missing key states by randomly removing a percentage 
of them, while Distraction simulates misidentification by adding distractor states through encouraging a randomly selected state dimension to 0.
Performance degrades shown in Table~\ref{tab:ksrobustness} with increasing perturbation, highlighting the importance of key state quality.
Nonetheless, LEMAE maintains strong robustness under both types of noise, ensuring reliable performance even under imperfect LLM outputs, which is valuable given the LLMs' limitations.

\begin{table}[t]
  \centering
  \caption{Robustness analysis of LEMAE to perturbations in key states, whether randomly deleting key states~(Reduction) or adding distracting states~(Distraction). }
  \resizebox{0.5\linewidth}{!}{
    \begin{tabular}{c|c|cc|cc}
    \toprule
          &  &\multicolumn{2}{c|}{Reduction} & \multicolumn{2}{c}{Distraction} \\
          \cline{3-6}
         Tasks & Default & \footnotesize{25\%} & \footnotesize{50\%} & \footnotesize{50\%}  &\footnotesize{100\%} \\
    \midrule
    1c3s5z & \textbf{0.98\scriptsize{$\pm$0.02}} & 0.97\scriptsize{$\pm$0.01} & 0.97\scriptsize{$\pm$0.02} & 0.92\scriptsize{$\pm$0.04} & 0.89\scriptsize{$\pm$0.05} \\
    3s\_vs\_5z & \textbf{0.83\scriptsize{$\pm$0.07}} & 0.80\scriptsize{$\pm$0.18} & 0.57\scriptsize{$\pm$0.28} & 0.80\scriptsize{$\pm$0.11} & 0.66\scriptsize{$\pm$0.08} \\
    MMM2 & \textbf{0.89\scriptsize{$\pm$0.08}} & \textbf{0.89\scriptsize{$\pm$0.03}} & 0.79\scriptsize{$\pm$0.09} &  0.86\scriptsize{$\pm$0.04} & 0.79\scriptsize{$\pm$0.08} \\
    \bottomrule
    \end{tabular}%
    }
  \label{tab:ksrobustness}%
\end{table}

\section{Further Discussions}\label{appsec:discussion}
\subsection{The Insights Behind Key States Discrimination}\label{discussion:discrimination}

In the considered scenarios, we claim that \textbf{key state discrimination} is generally more feasible and universally applicable than direct \textbf{key state generation} by LLM, especially under high-dimensional state spaces and partial observability. 
This reflects the widely held hypothesis in computer science regarding the \textit{generation-verification gap}—that verifying solutions is often significantly simpler than generating them~\cite{swamy2025roadsleadlikelihoodvalue, cook2023complexity}.
The rationale is threefold:
\begin{enumerate}
    \item 
    \textbf{Lower complexity and reduced error risk:} Discrimination focuses on high-level task understanding and identifying key state characteristics, while generation requires detailed, low-level comprehension, assigning values to each element. This imposes a significantly higher cognitive and computational load on the LLM, making generation more error-prone, particularly in high-dimensional settings. Discrimination equivalently simplifies the output space to key state labels, thus alleviating issues like hallucinations.
    \item \textbf{Greater transparency and debuggability:}  Errors in key state discrimination are easier to diagnose and fix, as the functions can be directly evaluated against actual states. In contrast, the quality of generated key states is often only indirectly observed through downstream performance (e.g., success rate), making errors harder to attribute and correct.

    \item \textbf{Robustness under partial observability:} 
    In cases of partial observability, generating key states directly is unreliable. For example, in the \textit{Pass} task, the positions of hidden switches are unknown and must be inferred from the door's status. LLM cannot generate key states accurately without knowledge of the specific agents' positions required to activate a switch. In contrast, it can still discriminate whether a given state indicates meaningful task progression by leveraging observable signals, such as whether the door is open.
\end{enumerate}

\subsection{Comparison with Options, Skills, and Bottleneck States}

To clarify the conceptual novelty of our approach, we distinguish Key States from two related concepts in RL: Options (or Skills) and Bottleneck States.

\textbf{Vs. Options and Skills.} 
Options and skills are \textit{temporal abstractions} representing action sequences or sub-policies (``how to reach''). 
In contrast, Key States are \textit{state abstractions} serving as static semantic subgoals (``what to reach''). 
LEMAE leverages Key States to guide a flat policy via intrinsic rewards (SHIR), without introducing additional hierarchical policy layers or separate sub-policy optimization.

\textbf{Vs. Bottleneck States.} 
Bottleneck states are typically defined \textit{topologically}, e.g., as structurally important connectors in the state-transition graph, and are often identified from visitation statistics or graph analysis after sufficient exploration. 
Key States, in contrast, are \textit{semantic} task milestones identified \textit{a priori} through LLM-based reasoning. 
They encode task logic and multi-agent coordination structure, which may not be explicitly captured by purely structural connectivity measures.

\subsection{Potential Extension to Incorporate Online LLM Feedback}

Our framework can naturally incorporate online LLM feedback as an extension of the existing self-check mechanism. Specifically, when a new candidate key state is identified during training, the LLM can be invoked to perform a self-check, reasoning about its correctness and task relevance before inserting it into the KSMT. If inconsistencies or weak justifications are detected, the key state discriminator can be refined accordingly.

This design does not require structural changes to LEMAE, but rather extends the current self-check procedure from offline validation to selective online verification. Such a mechanism would enable dynamic refinement of key states while keeping computational overhead controlled through conditional triggering.

\subsection{Potential Extension to Open-Environment and Embodied Multi-Agent System}

\textbf{Extension to Open-Environment MARL~\cite{yuan2023survey}:}
In open-environment settings, where agent populations, dynamics, or tasks may change over time, the core advantage of LEMAE lies in its semantic abstraction. Since key states represent high-level logical milestones rather than environment-specific low-level features, they can generalize across varying agent configurations or partially unseen scenarios. Moreover, the LLM prior can naturally incorporate updated task descriptions or environmental context, enabling dynamic re-localization of key states when new agents or objectives are introduced. This makes LEMAE particularly suitable for environments with evolving structures.

\textbf{Application to Embodied Multi-Agent Systems (MAS)~\cite{feng2025multi}:}
For embodied MAS, where agents operate in physically grounded environments with perception–action loops, key states can be grounded in semantic observations (e.g., object interactions, spatial relations, coordination events). LLM-informed key state localization can help identify meaningful coordination milestones, while KSMT provides structured guidance over long-horizon embodied tasks. Since LEMAE operates at the level of semantic state abstraction, it can be integrated with existing perception modules or foundation models without altering low-level control policies.

Overall, LEMAE offers a modular and extensible framework, where semantic key states serve as an interface between high-level reasoning and low-level multi-agent control, making it well-suited for open and embodied settings.

\section{Proof of Proposition~1}\label{appsec:lemmaproof}

\textbf{Proposition 1. }
Consider a one-dimensional asymmetric random walk starting at $0$ and targeting $N>1$, with right-move probability $p\in(0.5,1)$.
Without prior knowledge, the expected first hitting time is $\mathbb{E}(T_{0\rightarrow N}) = \frac{N}{2p-1}$.
After introducing the task-relevant information that the agent must first reach key states $\kappa=1, \dots, N-1$ before reaching $x=N$, the hitting time decreases $\mathbb{E}(T_{0\rightarrow N}) - \mathbb{E}(T^{\mathrm{prior}}_{0\rightarrow N}) = (N-1)*(\frac{1}{2p-1} - \frac{2}{p}+1) > 0$.  
\\

\begin{proof}
Random walk is a fundamental stochastic process, formed by successive summation of independent, identically distributed random variables~\cite{lawler2010random}.
This work considers the one-dimensional asymmetric random walk problem, where an agent starts at $x=0$ and aims to reach $x=N\in\mathbb{N^+}, N>1$. 
We use the expected first hitting time as a measure of performance, which corresponds to the average computational time complexity in this setting~\cite{yu2008new}.

\textbf{Step 1: Formulation of the Random Walk Process.}

Let $ M_0 = 0 $ denote the starting position, and $ M_1, M_2, \dots $ be independent and identically distributed (i.i.d.) random variables with the following distribution:

\begin{equation}
P(M_i = 1) = p, \quad P(M_i = -1) = 1 - p, \quad p \in (0.5, 1).
\end{equation}

The position of the agent after $ n $ steps is given by:

\begin{equation}
S_n = \sum_{i=1}^n M_i, \quad S_0 = 0.
\end{equation}

\textbf{Step 2: Martingale Construction.}

Due to the asymmetry of the random variables $ M_i $, the sequence $ \{S_n\}_{n \geq 0} $ is not a martingale with respect to $ \{M_n\}_{n \geq 1} $. However, the expected value of each random variable is:
\begin{equation}
\mathbb{E}(M_i) = 2p - 1 \quad \text{for all } i \geq 1.
\end{equation}

We introduce the process $ Y_n $ to account for the drift:
\begin{equation}
    Y_n = \sum_{i=1}^n\left(M_i-(2p-1)\right), \quad Y_0=0
\end{equation}

It is straightforward to verify that:
\begin{equation}
    \mathbb{E}|Y_n| = \sum_{i=1}^n \mathbb{E}|M_i| - n(2p - 1) = 2n - 2np < \infty.
\end{equation}
Additionally, for each $ n $, the conditional expectation of $ Y_{n+1} $ given $ M_0, M_1, \dots, M_n $ is:
\begin{equation}
    \mathbb{E}(Y_{n+1}|M_0,M_1,...M_n)=Y_n + \mathbb{E}(M_{n+1})-(2p-1) = Y_n
\end{equation}
Thus, $ \{Y_n\}_{n \geq 0} $ is a martingale with respect to $ \{M_n\}_{n \geq 1} $.

\textbf{Step 3: Stopping Time and Optional Stopping Theorem.}

Let $ T_{0 \rightarrow N} = \min\{n : S_n = N\} $ denote the first time the agent reaches $ x = N $. We can express it in terms of $ Y_n $:

\begin{equation}
T_{0 \rightarrow N} = \min\{n : Y_n = N - n(2p - 1)\}.
\end{equation}

It's clear that $T_{0\rightarrow N}$ is a stopping time with respect to $ \{M_n\}_{n \geq 1} $.

Next, we prove that:
\begin{equation}
    \mathbb{E} \left( |Y_{n+1} - Y_n| \mid M_0, M_1, \dots, M_n \right) = \mathbb{E} \left( |M_{n+1}| \right) - (2p - 1) = 2 - 2p < 2.
\end{equation}

We can assume that $\mathbb{E}(T_{0\rightarrow N})<\infty$. Thus, we can apply the Optional Stopping Theorem~\cite{durrett2019probability}, which gives:
\begin{equation}
    \mathbb{E}(Y_{T_{0\rightarrow N}}) = N-\mathbb{E}(T_{0\rightarrow N})*(2p-1)= \mathbb{E}(Y_0) = 0
\end{equation}
This results in:
\begin{equation}
    \mathbb{E}(T_{0\rightarrow N}) = \frac{N}{2p-1}
\end{equation}
The assumption $\mathbb{E}(T_{0\rightarrow N})<\infty$ is thereby validated. 

Consequently, the expected first hitting time within the default setting is:
\begin{equation}
\mathbb{E}(T_{0 \rightarrow N}) = \frac{N}{2p - 1},
\end{equation}
which agrees with the result stated in Theorem 4.8.9 of \cite{durrett2019probability}.

\textbf{Step 4: Incorporating Task-Relevant Information.}

We now introduce task-specific information that the agent must first reach intermediate key states, $ \kappa = 1, 2, \dots, N-1 $, before progressing to $ x = N $. Each time the agent reaches state $ x = \kappa $, the policy is updated to enforce deterministic rightward movement for all positions $ x < \kappa $, i.e., $ P(M_x = 1) = 1 $ for $ x < \kappa $, in line with the update process in Reinforcement Learning.

The expected first hitting time from $x=0$ to $x=1$ is:
\begin{equation}
\mathbb{E}(T_{0\rightarrow 1}) = \frac{1}{2p-1}
\end{equation} 

Next, the expected first hitting time from $x=1$ to $x=2$ is given by:
\begin{equation}
    \mathbb{E}(T^{\mathrm{prior}}_{1\rightarrow 2}) = p*\sum^\infty_{n=0}(2n+1)(1-p) = \frac{2}{p}-1
\end{equation}

Similarly, the expected first hitting time from $ x = 1 $ to $ x = N $ is:
\begin{equation}
    \mathbb{E}(T^{\mathrm{prior}}_{1\rightarrow N}) = (N-1)*\left(\frac{2}{p}-1\right)
\end{equation}

Consequently, the expected first hitting time after the integration of priors becomes 
\begin{equation}
\mathbb{E}(T^{\mathrm{prior}}_{0\rightarrow N}) =\mathbb{E}(T_{0\rightarrow 1})+\mathbb{E}(T^{\mathrm{prior}}_{1\rightarrow N}) = \frac{1}{2p-1} + (N-1)*\left(\frac{2}{p}-1\right)
\end{equation}

\textbf{Step 5: Total Advantage.}

The total advantage gained by incorporating priors is:
\begin{equation}
\mathbb{E}(T_{0\rightarrow N})- \mathbb{E}(T^{\mathrm{prior}}_{0\rightarrow N}) = (N-1)*(\frac{1}{2p-1} - \frac{2}{p}+1) > 0, p\in(0.5,1),N\in\mathbb{N}^+,N>1
\end{equation}

This completes the proof.

\end{proof}

\section{Implementation Details}\label{appsec:expdetails}

\subsection{LEMAE and Baselines}\label{appsec:baseline}

\textbf{LEMAE}: Our code is  based on the widely-used code framework pymarl2 at \url{https://github.com/hijkzzz/pymarl2}. In this study, we have integrated our method with several base algorithms \textbf{IPPO}, \textbf{QMIX}, \textbf{QPLEX}, and \textbf{VMIX}. Throughout the integration process, we have refrained from modifying the algorithmic implementation and have maintained consistency in parameters, including batch size, learning rate, and loss coefficients, in alignment with the configurations of the base algorithms.

\textbf{EITI} and \textbf{EDTI}~\cite{wang2019influence}: We compare our method with \textbf{EITI} and \textbf{EDTI} on MPE tasks proposed in CMAE~\cite{liu2021cooperative}. We use the experiment results reported in CMAE~\cite{liu2021cooperative}, which found that these algorithms perform poor because a long
rollout~(512 steps × 32 processes) between model updates
is used.

\textbf{CMAE}~\cite{liu2021cooperative}: We compare our method with \textbf{CMAE} on MPE and SMAC tasks. On MPE tasks,  the results of CMAE are reproduced using the publicly available code released by the authors at \url{https://github.com/IouJenLiu/CMAE}. As CMAE lacks an implementation for SMAC, we use the results reported in the original paper.

\textbf{MASER}~\cite{jeon2022maser}: 
Following the choice in LAIES~\cite{liu2023lazy}, we use the code in \url{https://github.com/Jiwonjeon9603/MASER}. Its suboptimal performance is also documented in LAIES.

\textbf{LAIES}~\cite{liu2023lazy}: 
We employed the publicly accessible code provided by the authors, which can be accessed at \url{https://github.com/liuboyin/LAIES}. When conducting experiments on SMAC, we adhered to the default configurations and external states. Notably, the original LAIES paper evaluation did not include assessments on the MPE. Consequently, we integrated the MPE environment into the LAIES codebase, designating the external states to represent the door status or the position of the box.

\textbf{ELLM}~\cite{du2023guiding}: 
Since the tasks in this work have clearly defined goals, we minimize LLM inference costs by following the ELLM methodology but adapting its goal generation to occur only once at the start of the training. Consistent with the hyperparameters in the official codebase~\url{https://github.com/yuqingd/ellm}, we set the similarity threshold to 0.99, rewarding only when the goal is achieved.
We rely on LLM-generated functions to verify goal achievement, which we found to be more effective than directly using semantic similarity-based rewards.

\textbf{ProgressCount~\cite{sarukkai2024automated}:} ProgressCount combines LLM-based reward design with count-based intrinsic rewards. We reimplement it following the original paper.

\textbf{FOX~\cite{jo2024fox} and ICES~\cite{li2024ices}:} FoX proposes a formation-based equivalence abstraction to reduce the search space, encouraging agents to explore diverse formations under partial observability. ICES designs agent-wise intrinsic rewards based on Bayesian surprise in global latent transitions. We implement FoX and ICES based on their official codebases and recommended settings.

For all algorithms, we ensure the same environmental settings, including observation space, environment reward function, and so on. 

For all LLM-based methods, we provide the same task description and state form as input, without access to additional information. 
Concretely, for SMAC, MPE, and MaMuJoCo, task descriptions and state/observation definitions are obtained from their respective official documentations and benchmark papers\footnote{\url{https://github.com/oxwhirl/smac/blob/master/docs/smac.md}},\footnote{\url{https://arxiv.org/pdf/2107.11444}},\footnote{\url{https://gymnasium.farama.org/environments/mujoco}}.
Method-specific instructions are applied only as required by each baseline. This ensures a fair comparison, with observed performance differences reflecting the method itself rather than variations in available information.

\subsection{Comparison with LLM Reward Design}\label{appsec:llmrd}
We conduct additional experiments comparing LEMAE with a baseline called \textbf{Eureka-si}, which can be seen as a single-iteration variant of Eureka~\cite{ma2023eureka}, where LLM designs rewards directly.  For fairness, we does not adopt evolutionary optimization in Eureka and use LLM to generate reward functions with the same role instructions as in Eureka, while maintaining designs like Self-Check as in LEMAE. 
As shown in Figure~\ref{fig:roomresult}, Eureka-si is comparable to LEMAE in simple tasks like Push-Box but fails in challenging tasks with characteristics like partial observability, such as Pass, where hidden switches make it difficult to design effective reward functions. In contrast, LEMAE consistently demonstrates impressive performance. Notably, comparing LEMAE with Eureka directly would be unfair since Eureka's evolutionary search requires multiple training iterations and candidates, leading to significantly more sampling and training than LEMAE. 
Overall, LEMAE's advantage over RL algorithms lies in incorporating prior knowledge from the LLM, and its advantage over other LLM-based methods is due to our designs for better LLM incorporation, such as utilizing discrimination, SHIR, and KSMT.

\subsection{Connection and Comparison with HER}\label{appsec:her}

The proposed Key State-Guided Exploration is similar to Hindsight Experience Replay (\textbf{HER})~\cite{andrychowicz2017hindsight} in form, where key states and subgoals are certain states from sampled trajectories. 
However, unlike HER, which samples goals from memory using random or heuristic strategies and often struggles with shaped rewards, our method incorporates LLM priors for more targeted goal selection (key states localization). Additionally, the proposed KSMT and SHIR facilitate organized exploration and enhanced reward guidance. 

We conduct additional experiments to further confirm the advantages of our method.
We evaluate HER with IPPO as the backbone in MPE. We use the future strategy for goal selection, as proposed in the HER paper, and employ a reward function based on the Manhattan Distance, which we find to be the best match. However, as depicted in Figure~\ref{fig:roomresult}, HER does not perform well on all MPE tasks. This outcome suggests that the random sampling strategy for goals may not be sufficient, underscoring the importance of incorporating LLM priors for efficient exploration as we proposed.

\subsection{Tasks}\label{appsec:envdetails}

\subsubsection{Multiple-Particle Environment~(MPE)}
The Multiple-Particle Environment serves as a widely-adopted benchmark for multi-agent scenarios. In this work, we employ tasks specifically crafted for evaluating multi-agent exploration, proposed by \cite{wang2019influence}. The implementation utilized in this study is based on the work by \cite{liu2021cooperative}. In this section, we provide details of the four sparse-reward tasks we adopted.

$\bullet$ \textit{Pass}: 
In the \textit{Pass} task, depicted in Figure~1(c)(i), two agents are positioned in a room of 30 x 30 grid. The room is divided into two halves by a wall featuring a door. Each half-room contains an invisible switch, the details of which are not contained in the state or prompt for LLM. The door permits passage only when one of the switches is occupied by an agent. Initially situated within the left half-room, both agents must cooperate to transfer to the right half-room. The external reward function is denoted as $r_E=I(two\ agents\ are\ in\ the\ right\ room)$, where $I$ represents  the indicator function.

$\bullet$ \textit{Secret-Room}: 
\textit{Secret-Room} is an extension task of \textit{Pass}. As illustrated in Figure~1(c)(ii), the configuration comprises one sizable room on the left and three smaller rooms on the right, interconnected by three doors. Within each room, there is an invisible switch; notably, the switch in the left room has the capability to control all three doors, whereas each right room's switch exclusively controls its respective door. The grid size is 25 x 25. Two agents are initialized within the left room and are required to collaborate in order to transition to the real target room, which is the right room 2. The external reward function is denoted as $r_E=I(two\ agents\ are\ in\ the\ right\ room\ 2)$, where $I$ represents  the indicator function.

$\bullet$ \textit{Push-Box}: 
As depicted in Figure~1(c)(iii), two agents and a box are initially positioned within a 15 x 15 grid. To successfully move the box, both agents must simultaneously exert force in the same direction. The task is deemed accomplished when the box is successfully pushed to the wall. The external reward function is denoted as $r_E=I(the\ box\ is\ pushed\ to\ the\ wall)$, where $I$ represents  the indicator function.

$\bullet$ \textit{Large-Pass}: 
\textit{Large-Pass} is a direct extension task of \textit{Pass} by enlarging the grid dimensions to 50 x 50, which makes it more challenging. The external reward function aligns with that of the \textit{Pass} task.

The details of these tasks, including observation space and action space, are listed in Table~\ref{apptab:mpe_tasks}.

\begin{table}[htbp]
\centering
\caption{Details of MPE tasks}
\resizebox{0.5\linewidth}{!}{
\begin{tabular}{ccccc}
\toprule
\textbf{MPE tasks} & \textbf{n\_agents} & \textbf{observation space} & \textbf{state space} & \textbf{action space} \\ \midrule
Pass                & 2                   & 5                           & 5                     & 4                     \\ 
Secret-Room         & 2                   & 5                           & 5                     & 4                     \\ 
Push-Box            & 2                   & 6                           & 6                     & 4                     \\ 
Large-Pass          & 2                   & 5                           & 5                     & 4                     \\ \bottomrule
\end{tabular}
}
\label{apptab:mpe_tasks}
\end{table}

\subsubsection{StarCraftII Multi-Agent Challenge~(SMAC) v1 and v2}
StarCraftII Multi-Agent Challenge~(SMAC) v1~\cite{samvelyan19smac} and v2~\cite{ellis2024smacv2} is a widely-used benchmark in the realm of cooperative multi-agent reinforcement learning research~\cite{rashid2018qmix,shao2023complementary,liu2023lazy, shao2024counterfactual}. 
Derived from the renowned real-time strategy game StarCraft II, SMAC concentrates specifically on decentralized micromanagement scenarios rather than the full game.
Typically, the tasks within SMAC adopt a dense-reward framework, wherein agents receive dense rewards for damage received, attacking and eliminating enemies. 
To promote the need for exploration, we adopt fully sparse-reward versions of tasks in SMAC where agents are solely rewarded upon the successful elimination of all enemies. 
The external reward function is denoted as $r_E=I(all\ enemies\ are\ eliminated )$, where $I$ represents  the indicator function.
Notably, this sparse-reward setting differs from the sparse SMAC, which can be called semi-sparse SMAC, used in some previous studies~\cite{jeon2022maser, jo2024fox}, where agents are rewarded when one or all enemies die or when one ally dies. 
In addition, to validate the versatility of LEMAE across diverse scenarios, we conducted experiments on ten maps with different difficulty and diverse agent numbers, as illustrated in 
Table~\ref{apptab:smac_taskdetails}.
Please refer to the official document\footnote{\url{https://github.com/oxwhirl/smac/blob/master/docs/smac.md}}\footnote{\url{https://github.com/oxwhirl/smacv2}} for more details.

\begin{table}[t]
  \centering
  \caption{Details of SMAC tasks}
  \resizebox{0.7\linewidth}{!}{
    \begin{tabular}{ccccccc}
    \toprule
    \textbf{Version}& \textbf{SMAC tasks} & \textbf{n\_agents} &\textbf{n\_enemies} & \textbf{observation space} & \textbf{state space} & \textbf{action space} \\ \midrule
    \multirow{5}{*}{SMACv1}&2m\_vs\_1z & 2 & 1 & 16 & 26 & 7 \\
    &1c3s5z & 9 & 9 & 162 & 270 & 15 \\
    &3s\_vs\_5z & 3 & 5 & 48 & 68 & 11 \\
    &3s5z\_vs\_3s6z & 8 & 9 & 136 & 230 & 15 \\
    &MMM2 & 10 & 12 & 176 & 322 & 18 \\
    \midrule
    \multirow{5}{*}{SMACv2}&protoss\_5\_vs\_5 & 5 & 5 & 92 & 130 & 11\\
    &terran\_5\_vs\_5 & 5 & 5 & 82 & 120 & 11\\
    &zerg\_5\_vs\_5 & 5 & 5 & 82 & 120 & 11\\
    &terran\_10\_vs\_11 & 10 & 11 & 170 & 306 & 17\\
    &zerg\_10\_vs\_11 & 10 & 11 & 170 & 306 & 17\\
    \bottomrule
    \end{tabular}%
    }
  \label{apptab:smac_taskdetails}%
\end{table}%

\subsection{Hyperparameters}\label{appsec:hyperparameter}

In LEMAE, we introduce three important hyperparameters: extrinsic reward scaling rate $\alpha$, intrinsic reward scaling rate $\beta$, and high randomness epsilon $\epsilon_h$. Notably, the low randomness epsilon $\epsilon_l$ is the hyperparameter in the base algorithms, such as 0.05 for QMIX and 0.0 for IPPO.

For MPE, we adopt $\{\alpha=10, \beta=0.1, \epsilon_h=1\}$ on \textit{Pass}, \textit{Secret-Room}, and \textit{Large-Pass} and use $\{\alpha=10, \beta=0.05, \epsilon_h=0.2\}$ on \textit{Push-Box}.

For SMAC, we adopt $\{\alpha=10, \beta=1, \epsilon_h=0.5\}$ on \textit{3s\_vs\_5z}, \textit{2m\_vs\_1z} and all SMACv2 tasks, $\{\alpha=50, \beta=1, \epsilon_h=0.5\}$ on \textit{MMM2} and \textit{1c3s5z}, $\{\alpha=1, \beta=1, \epsilon_h=0.5\}$ on \textit{3s5z\_vs\_3s6z}.

\section{Scalability and Generalization Analysis}\label{appsec:scale}

\subsection{A Brand New Task: \textit{River}}
\label{appsec:river}
\begin{figure*}[t]
\centering

\begin{minipage}[t]{0.39\linewidth}
    \centering
    \includegraphics[width=\linewidth]{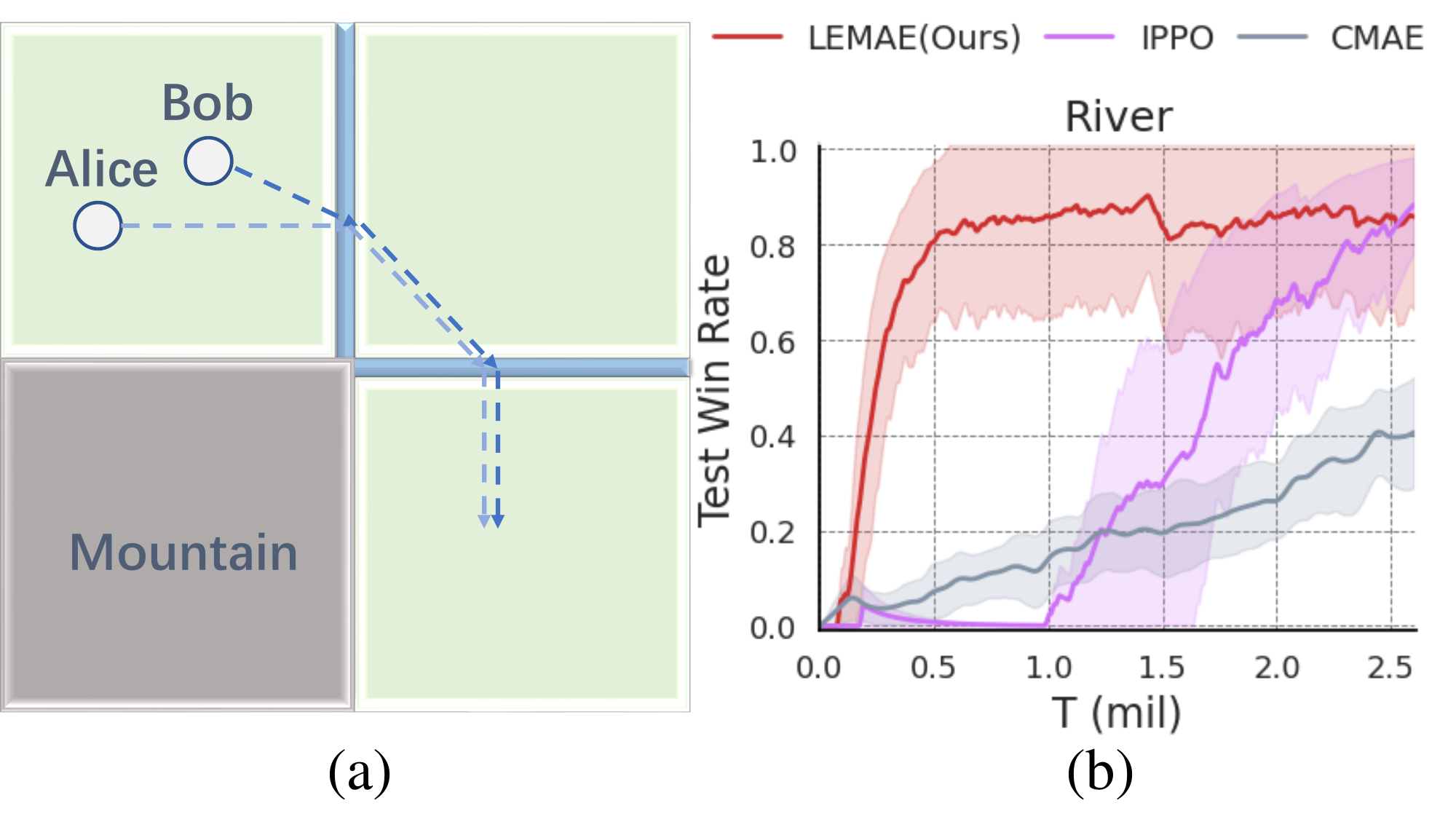}
    \vspace{-15pt}
    \caption{A brand new task, \textit{River}, which LLM has never encountered before, together with the training curves of LEMAE and baselines using the evaluation metric of test success rate.}
    \label{fig:river}
\end{minipage}
\hspace{2pt}
\begin{minipage}[t]{0.59\linewidth}
    \centering
    \includegraphics[width=\linewidth]{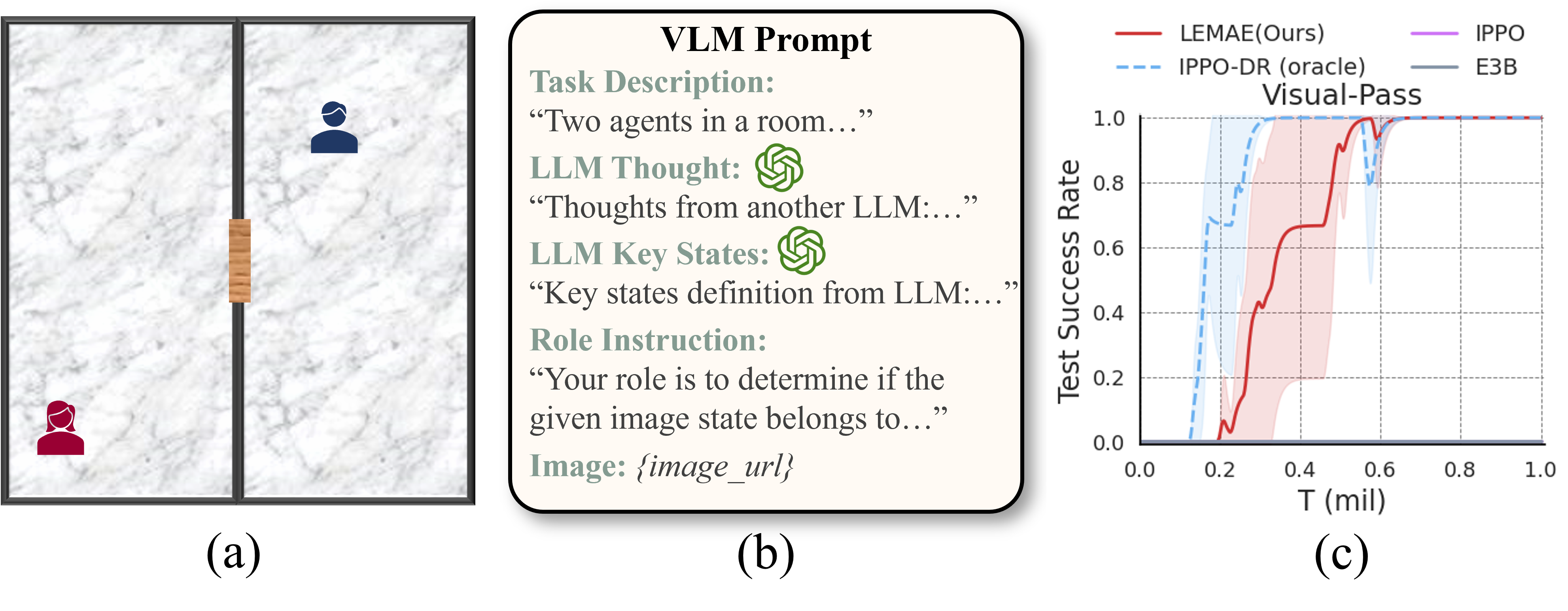}
    \vspace{-15pt}
    \caption{An illustration of the vision-based task \textit{Visual-Pass}. The prompt for the Vision Language Model~(VLM) includes the task description, role instruction, image state and key states definition provided by another LLM. The VLM is tasked with determining whether the given image state corresponds to a key state. Training curves of LEMAE on the image-based \textit{Visual-Pass} tasks, where \textbf{IPPO-DR} refers to IPPO trained with human-designed dense rewards.}
    \label{appfig:vlm}
\end{minipage}

\vspace{-5pt}

\end{figure*}

To exclude the probability that LEMAE's success relies on LLM's familiarity with the chosen tasks, we've designed a brand new task, termed \textit{River}, which \textbf{LLM has never encountered before}. The task is detailed as follows:

The \textit{River} task is adapted from MPE and its map is illustrated in Figure~\ref{fig:river}(a). 
Two agents, Alice and Bob, are placed in a 30 x 30 grid field intersected by two rivers running vertically and horizontally. A mountain in the bottom-left corner obstructs the passage. Alice and Bob start randomly in the top-left part of the field and need to move to the bottom-right part. However, Alice is afraid of water and cannot cross the river unless Bob stays in the river to act as a bridge for her.

The observation space is discrete with four dimensions, representing the positions of two agents, i.e., $o=[x_1,y_1,x_2,y_2]$. The action space is also discrete, allowing movement in four directions. Agents receive a positive reward only when both agents reach the bottom-right corner of the field.

As shown in Figure~\ref{fig:river}(b), LEMAE outperforms the baselines, and this confirms LLM's generalization capabilities to empower LEMAE's effectiveness in promoting efficient exploration in diverse new tasks.

\subsection{Extending LEMAE Beyond Symbolic Tasks}\label{appsec:imgbased}

This work primarily focuses on tasks with symbolic state spaces, where states are represented as symbolic arrays describing the agent and environment.
To extend LEMAE from symbolic tasks to vision-based tasks, we can exchange the LLM for a multi-modal LM in LEMAE for key state localization. 

To confirm the applicability of LEMAE to vision-based tasks, we conduct a demonstrative experiment: We extend the task \textit{Pass} to a vision-based task \textit{Visual-Pass}, as illustrated in Figure~\ref{appfig:vlm}(a). 
We prompt a LLM to define key states with the same task description and role instruction as proposed in 4 and use the LLM-generated definition as the prompt for a \textbf{Vision Language Model (GPT-4o)}. 
Then, it is prompted to discriminate key states in the randomly sampled states. 
GPT-4o achieves a \textbf{98\% accuracy rate in discriminating key states} among the 50 sampled image states. 
This confirms that with a proper extension of the LLM, LEMAE can eliminate dependence on state semantics and be applied to other tasks such as visual input.

As shown in Figure~\ref{appfig:vlm}c, we evaluate the effectiveness of LEMAE in the image-based task \textit{Visual-Pass} by using a VLM as the discriminator and the detection of pixel objects for the calculation of the reward. 
LEMAE demonstrates superior performance over the baseline IPPO and a generic intrinsic reward method E3B~\cite{E3B}, achieving comparable results to IPPO-DR, an oracle baseline with dense rewards.
The experimental results show the broader applicability of LEMAE.

\subsection{Experiments for Single-Agent setups}\label{appsec:single-agent}

LEMAE holds the promise to be a general approach for LLM-empowered efficient exploration in reinforcement learning, applicable to both single-agent and multi-agent settings. We underscore the evaluation of its performance in multi-agent settings due to its inherent complexity. 

As the proposed method can be easily extended to single-agent scenarios, we introduce a single-agent variant of MPE and assess PPO~\cite{schulman2017proximal} and PPO-based LEMAE for four tasks. We run each algorithm using three random seeds with 300k environment steps, using the evaluation metric of the test success rate. The following table shows that LEMAE can facilitate efficient exploration in single-agent scenarios.

\begin{figure*}[t]
\centering

\begin{minipage}[t]{0.36\linewidth}
\centering
\vspace{-100pt}
\captionof{table}{Final test success rate of LEMAE and PPO on single-agent variant of MPE tasks.}
\label{apptab:singleagent}

\resizebox{\linewidth}{!}{
\begin{tabular}{ccc}
\toprule
\textbf{Single MPE} & \textbf{PPO}& \textbf{LEMAE} \\
\midrule
Single Pass & 0.00\scriptsize{$\pm$0.00} & \textbf{1.00\scriptsize{$\pm$0.00}} \\
Single Secret-Room & 0.00\scriptsize{$\pm$0.00} & \textbf{0.98\scriptsize{$\pm$0.01}} \\
Single Large-Pass & 0.00\scriptsize{$\pm$0.00} & \textbf{0.99\scriptsize{$\pm$0.01}} \\
Single Push-Box & 0.00\scriptsize{$\pm$0.00} & \textbf{0.96\scriptsize{$\pm$0.08}} \\
\bottomrule
\end{tabular}
}

\end{minipage}
\hfill
\begin{minipage}[t]{0.6\linewidth}
\centering

\includegraphics[width=0.8\linewidth]{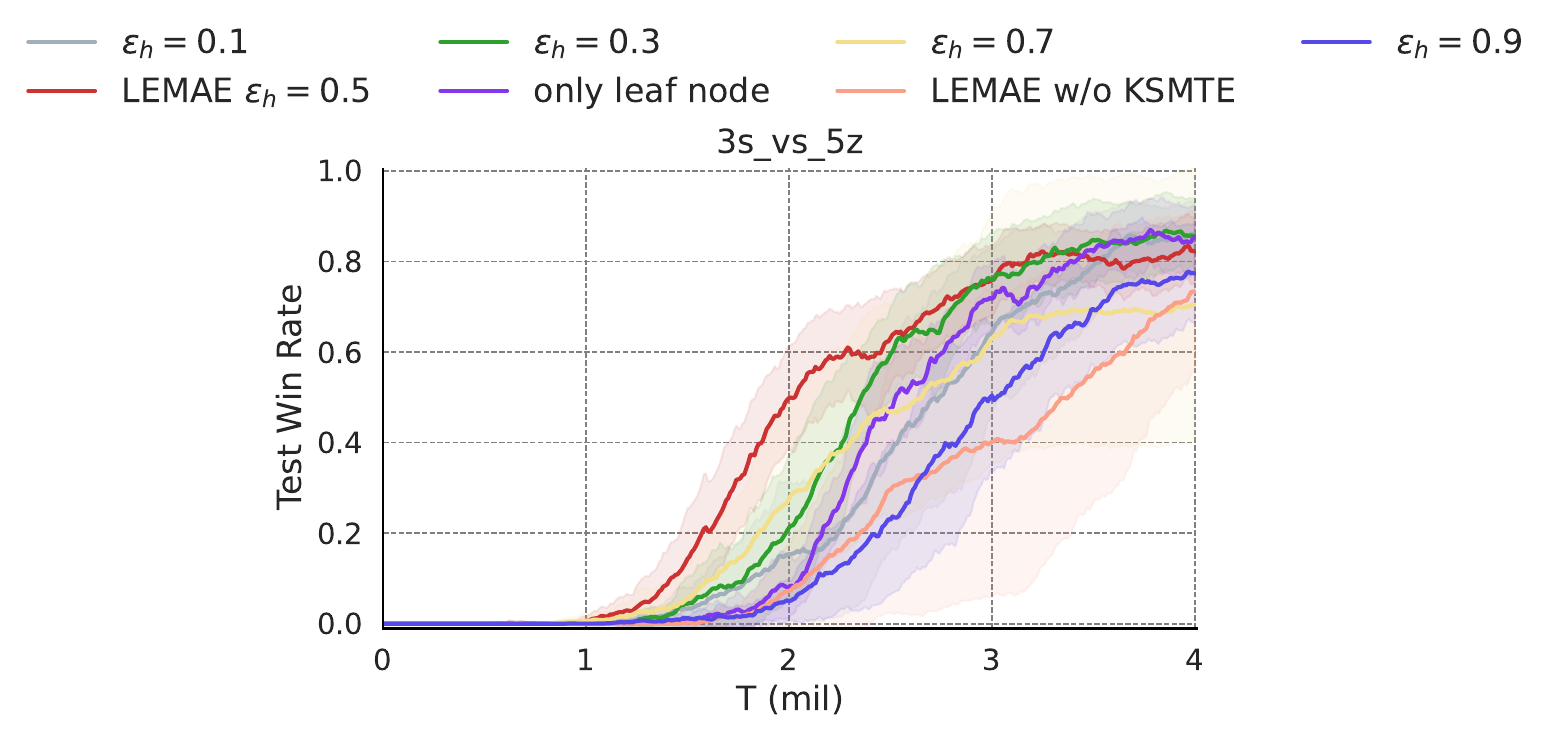}

\vspace{-10pt}

\caption{
Ablation study on mixed-randomness exploration. The default hyperparameter in LEMAE is $\epsilon_h=0.5$. 
Robustness is evaluated by varying $\epsilon_h$ across $\{0.1, 0.3, 0.7, 0.9\}$. “only leaf node” refers to applying $\epsilon_l$ at non-leaf nodes and $\epsilon_h$ at leaf nodes. “LEMAE w/o KSMTE” denotes a version of LEMAE without the mixed-randomness exploration module.
}

\vspace{-5pt}

\label{appfig:hybrid}

\end{minipage}

\end{figure*}

\section{Additional Experimental Results}
\label{appsec:addresults}
\subsection{Ablation Studies on Mixed-Randomness Exploration}\label{appsec:mixrand}

As shown in Figure~\ref{appfig:hybrid}, an ablation study is conducted on the mixed-randomness exploration strategy within the \textit{3s\_vs\_5z} map.
Specifically, LEMAE is evaluated under different values of the parameter $\epsilon_h$, which controls the degree of stochasticity in the high-randomness policy (i.e., an $\epsilon_h$-greedy policy).
The results demonstrate that LEMAE is robust to the choice of $\epsilon_h$ as long as the level of randomness remains moderate, avoiding extreme values such as 0.1 or 0.9. 
Furthermore, the effectiveness of the proposed strategy is corroborated by comparisons with two variants: a leaf-node-only version, which applies $\epsilon_l$ at non-leaf nodes and $\epsilon_h$ at leaf nodes, and a version of LEMAE without KSMTE, which disables the mixed-randomness exploration mechanism.

\subsection{Discussion on the KSMT}

Using KSMT could pose a limitation due to potential memory costs in certain scenarios. 
However, this has not been a significant issue in our experiments, as the key states are relatively few, primarily focusing on the most critical ones, with a natural sequential relationship typically existing between them.
Notably, LEMAE is also compatible with other memory structures, such as Directed Acyclic Graphs~(DAGs), which could be an interesting direction for future exploration.

To demonstrate the effectiveness of LEMAE with other memory structures, in scenarios where task completion follows a linear pattern (e.g., $Init \rightarrow A \rightarrow B \rightarrow Success$), we employ a more efficient strategy by using a KSMT variant with a single branch representing the sequential order of key states.
Specifically, we systematically assign a priority value to each key state, continuously updating it based on its occurrence order within the sequence of attained key states. The determination of the ranking of key states within the one-branch KSMT relies on this established priority.

\begin{figure}[t]
  \centering
    \includegraphics[width=\linewidth]{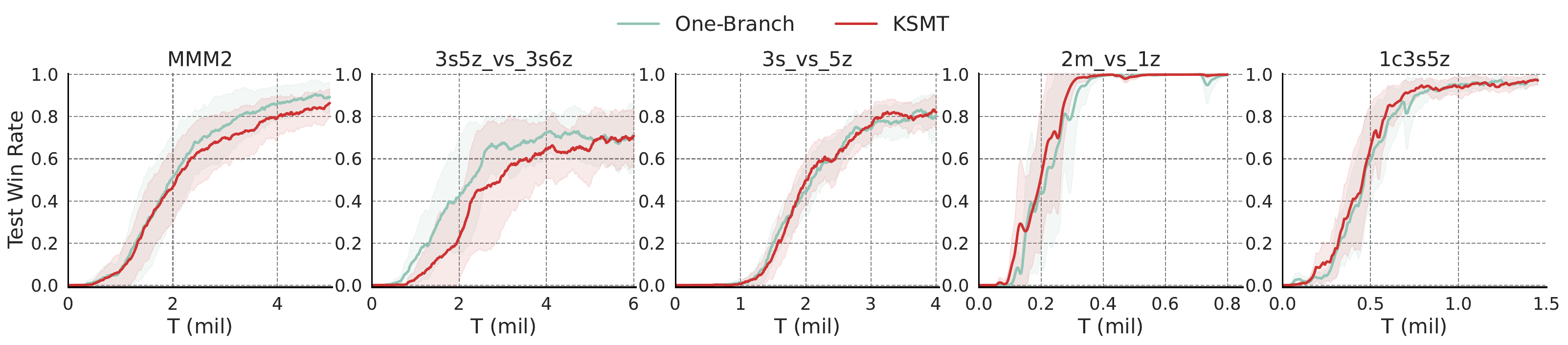}
    \vspace{-20pt}
    \caption{
Ablation study conducted to compare the performance between raw KSMT and the one-branch KSMT variant across six maps in SMAC.}
\vspace{-5pt}
    \label{appfig:one-branch}
\end{figure}

As illustrated in Figure~\ref{appfig:one-branch}, an ablation study is conducted to compare the performance between raw KSMT and the one-branch KSMT variant across six maps in SMAC. The results demonstrate the increased necessity of employing the one-branch KSMT variant for tasks involving a larger number of agents and greater complexity, such as \textit{3s5z\_vs\_3s6z} and \textit{MMM2}. Consequently, we have adopted the one-branch KSMT approach for these specific SMAC tasks: \textit{3s5z\_vs\_3s6z} and \textit{MMM2}.

\subsubsection{Connections with Go-Explore}
Go-Explore~\cite{ecoffet2019go} is an influential work tackling exploration in RL. The similarities between our KSMT and the archive in Go-Explore lie in both methods organizing exploration through memory, i.e., by selecting possible historical states to explore.
The differences and partial contributions of LEMAE are as follows: (1) Our key states are semantically meaningful and task-critical, whereas the archived states in Go-Explore are randomly explored; (2) Our KSMT samples key states based on actual key states transitions, enhancing its reliability; (3) We propose Explore with KSMT to balance exploration and exploitation, thereby reducing exploration complexity by focusing on a more meaningful state subspace.

\subsubsection{Complexity Analysis}

The KSMT is designed to be lightweight.
Key states serve as high-level, abstract guidance, and their total number is naturally small and sparse.
Appendix J shows that most scenarios have fewer than 20 key states.
Furthermore, logical transition relationships among key states result in a highly organized tree structure rather than a random or fully branched one.
These factors bound both the depth and width of KSMT, keeping construction and query operations computationally inexpensive.

To provide empirical evidence, additional experiments were conducted on typical SMAC. Results show that KSMT accounts for $\bm{\le6\%}$ of total computation time while substantially improving sample efficiency. These findings demonstrate that the memory tree introduces minimal overhead, even for long-horizon or multi-agent tasks, while offering significant gains in exploration performance. 

\subsubsection{Discussion on Branch Pruning.}
In the current implementation of KSMT, once a valid success state is discovered, the corresponding successful branch is retained while alternative branches are pruned. In principle, this strategy may discard trajectories that could potentially lead to globally optimal solutions. 
However, this design is motivated by the characteristics of sparse-reward cooperative environments considered in this work. Discovering any successful trajectory is often already challenging, and LLM-localized key states may introduce semantically plausible but task-irrelevant branches due to incomplete prior information. Preserving all branches can therefore lead to semantic branch explosion and significantly dilute exploration over effective directions.

The pruning strategy prioritizes empirically validated successful paths to stabilize and accelerate exploration. Meanwhile, since the intrinsic reward in LEMAE is progress-based and defined with respect to key-state completion rather than to a specific trajectory instance, policy optimization continues after success discovery and can still improve within the retained key-state structure.
Future extensions may explore more flexible branch management strategies, such as retaining multiple promising branches or periodically revisiting pruned ones, to further balance exploration efficiency and global optimality.

\section{LLM Prompts and Responses}\label{appsec:prompt}
Here are the example prompt and response in our work. Please reference the code for further details. Notably, we adopt the chain-of-thought technique from \cite{wei2022chain}.

\begin{tcolorbox}[title = {SMAC Prompt and Response Example}]
\textbf{SYSTEM:}\\

\textbf{(Task\_Description)}\\
We are playing StarCraft II micro scenario, tring to control our agents to defeat all of the enemy units.\\

\textbf{(State\_Form)}\\
In each step, the current state is represented as a 1-dimensional list: \\
$[nf\_al]*n\_agents + [nf\_en]*n\_enemies + [last\_actions]$.\\

$nf\_al$ denotes the unit state for each agent with attributes\\
$[health\_rate, weapon\_cooldown\_rate, relative\_x\_to\_map\_center,\\ relative\_y\_to\_map\_center, shield\_rate$ (1 dimension if a\_race is P else 0 dimension),\\ $unit\_type\_bits$ (the dimension is defined in the map config)$]$.\\

$nf\_en$ represents the unit state for each enemy with attributes\\ $[health\_rate, relative\_x\_to\_map\_center, relative\_y\_to\_map\_center, \\shield\_rate$ (1 dimension if b\_race in map config is P else 0 dimension), \\
$unit\_type\_bits$ (the dimension is defined in the map config)$]$. \\

The $last\_actions$ component does not require consideration.\\

\textbf{(Role\_Instruction)(Template)}\\
Your role is to give several critical key states in the task which we should try to reach and generate the corresponding discriminator function for each key state 
which can discriminate if the input state has reached the key state. \\
Note:\\
\hbox{\ \ \ \ } 1. Don't use the information you are not told. \\
\hbox{\ \ \ \ } 2. The code should be as generic as possible. \\
\hbox{\ \ \ \ } 3. The discriminator functions for different key states should be independent. \\
\hbox{\ \ \ \ } 4. Your answer should be complete and not omitted. \\

Please think step by step and adhere to the following JSON format (just replace the () with your answer):\\
\{\\
\hbox{\ \ \ \ } Thought: (Please fully understand the task. Think step by step and give your thought about the
concrete state here),\\
\hbox{\ \ \ \ } Key states: \{\\
\hbox{\ \ \ \ \ \ \ \ } init: all agents and enemies are alive, \\
\hbox{\ \ \ \ \ \ \ \ }  key\_state\_1: (description), \\
\hbox{\ \ \ \ \ \ \ \ } key\_state\_2: (description),\\
\hbox{\ \ \ \ \ \ \ \ } ..., \\
\hbox{\ \ \ \ \ \ \ \ } success: all enemies are dead and at least one agent is alive\\
\hbox{\ \ \ \ } \}, \\
\end{tcolorbox}
\begin{tcolorbox}
$\quad$ Discriminators: \{\\
\hbox{\ \ \ \ \ \ \ \ } key\_state\_1: (a python function with the form of def iskeystate1(state): ... return 0 or 1), \\
\hbox{\ \ \ \ \ \ \ \ } key\_state\_2: (a python function with the form of def iskeystate2(state): ... return 0 or 1),\\
\hbox{\ \ \ \ \ \ \ \ } ...\\
\hbox{\ \ \ \ } \}\\
\hbox{\ \ \ \ } Subspace\_states: \{\\
\hbox{\ \ \ \ \ \ \ \ } key\_state\_1: (extract subspace state used in corresponding discriminator functions, such as [1] or [2,5]), \\
\hbox{\ \ \ \ \ \ \ \ } key\_state\_2: (extract subspace state used in corresponding discriminator functions, such as [1] or [2,5]),\\
\hbox{\ \ \ \ \ \ \ \ } ...\\
\hbox{\ \ \ \ } \}\\
\}
\\

\textbf{USER:}\\
Task is $\textit{2m\_vs\_1z}$. 
The map config is: \{\\
\hbox{\ \ \ \ }n\_agents: 2, n\_enemies: 1, limit: 150, a\_race: T, b\_race: P, unit\_type\_bits: 0, map\_type: marines\\
\}. \\

Give more than 2 key states and the corresponding discriminator functions.
\\

\textbf{ASSISTANT: (LLM response)}\\
\{\\
\hbox{\ \ \ \ }Thought: Since a\_race is T, there is no shield\_rate in nf\_al; b\_race is P, so there is 1-dim shield\_rate in nf\_en; the dim of unit\_type\_bits is 0. The dimension of nf\_al = 4+0+0 = 4, the dimension of nf\_en = 3+1+0 = 4. The concrete state form is $[$agent\_0\_health\_rate, agent\_0\_weapon\_cooldown\_rate, agent\_0\_relative\_x\_to\_map\_center, agent\_0\_relative\_y\_to\_map\_center, agent\_1\_health\_rate, agent\_1\_weapon\_cooldown\_rate, agent\_1\_relative\_x\_to\_map\_center, agent\_1\_relative\_y\_to\_map\_center, enemy\_0\_health\_rate, enemy\_0\_relative\_x\_to\_map\_center, enemy\_0\_relative\_y\_to\_map\_center, enemy\_0\_shield\_rate, last\_actions$]$. Key states should focus on the health and shield status of the enemy and the health of our agents.\\
\hbox{\ \ \ \ }Key\_states: \{\\
\hbox{\ \ \ \ \ \ \ \ }init:all agents and enemies are alive,\\
\hbox{\ \ \ \ \ \ \ \ }key\_state\_1: enemy\_0's shield is significantly damaged,\\
\hbox{\ \ \ \ \ \ \ \ }key\_state\_2: enemy\_0 is significantly damaged,\\
\hbox{\ \ \ \ \ \ \ \ }key\_state\_3: enemy\_0 is about to die,\\
\hbox{\ \ \ \ \ \ \ \ }success:all enemies are dead and at least one agent is alive\\
\hbox{\ \ \ \ \ \ \ \ }\}, \\
\hbox{\ \ \ \ }Discriminators: \{\\
\hbox{\ \ \ \ \ \ \ \ }key\_state\_1: def iskeystate1(state):   return 1 if state[11] \textless 0.6 else 0,\\
\hbox{\ \ \ \ \ \ \ \ }key\_state\_2: def iskeystate2(state):   return 1 if state[8] \textless 0.6 else 0,\\
\hbox{\ \ \ \ \ \ \ \ }key\_state\_3: def iskeystate3(state):   return 1 if state[8] \textless 0.05 else 0,\\
\hbox{\ \ \ \ \ \ \ \ }\}\\
\hbox{\ \ \ \ }Subspace\_states: \{\\
\hbox{\ \ \ \ \ \ \ \ }key\_state\_1: [11],\\
\hbox{\ \ \ \ \ \ \ \ }key\_state\_2: [8],\\
\hbox{\ \ \ \ \ \ \ \ }key\_state\_3: [8],\\
\hbox{\ \ \ \ \ \ \ \ }\}\\
\}

\textbf{USER:}\\
\textbf{(LLM Rethinking)}\\
Please carefully check your response to make sure it meets the task requirements and notes without including unnecessary details. Also, confirm that the discriminator functions do not use any undefined variables.\\

\textbf{ASSISTANT: (LLM response)}\\
\textbf{......} (We have omitted the intermediate LLM outputs to maintain conciseness, retaining only the initial generation, as the differences between the initial and rechecked generations are minimal in the absence of errors.)
\end{tcolorbox}

\begin{figure*}[ht]
  \centering
    \includegraphics[width=0.5\linewidth]{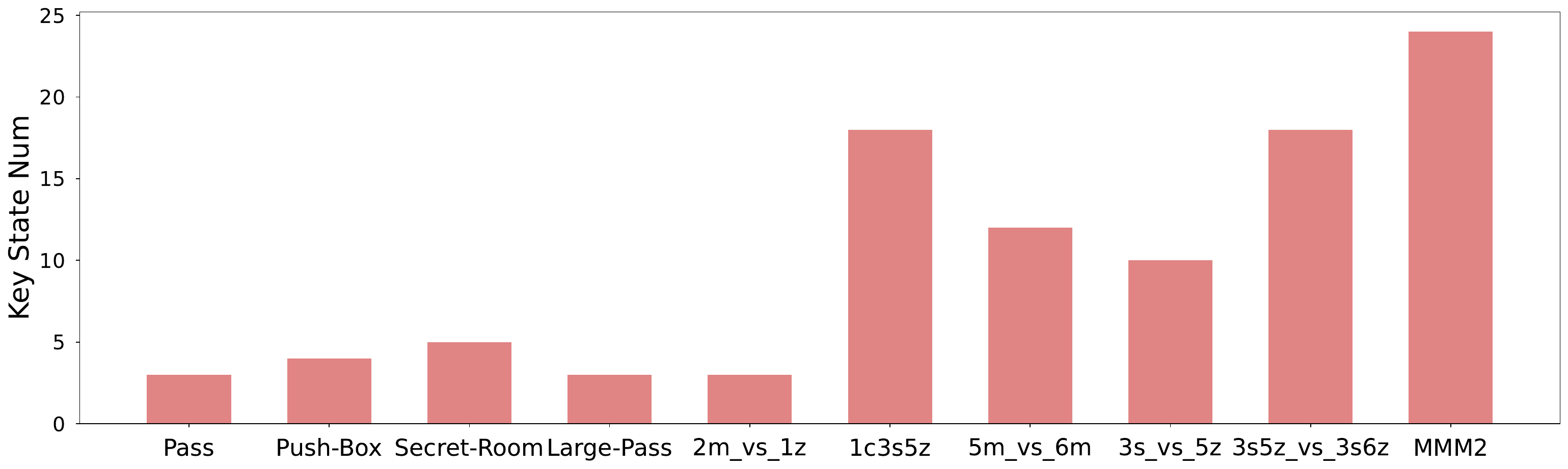}
    \vspace{-5pt}
    \caption{The number of LLM-generated key state discriminator functions.}
    \vspace{-5pt}
    \label{appfig:finalksn}
\end{figure*}

Notably, the number of key states is primarily determined by LLM. For each task, we only prompt LLM to prevent it from generating too few functions according to the complexity of the environment. Specifically, we instruct LLM to generate several critical key states for MPE and more than 2*n\_enemies critical key states for SMAC. 

As shown in Figure~\ref{appfig:finalksn}, we summarize the number of LLM-generated key state discriminator functions. It is notable that the number of discriminator functions increases with the difficulty of the task or the number of interactive objects in the environment, which aligns with intuition.
Additionally, we have omitted the intermediate LLM outputs to maintain conciseness in this section, retaining only the initial generation, as the differences between the initial and rechecked generations are minimal in the absence of errors.

\end{appendix}
\end{document}